\documentclass{article}

\usepackage{natbib}
\usepackage{geometry}
\usepackage{listings}
\usepackage{times}
\usepackage{latexsym}
\usepackage{amsmath}
\usepackage[T1]{fontenc}
\usepackage{amsfonts}
\usepackage[utf8]{inputenc}
\usepackage{microtype}
\usepackage{booktabs}
\usepackage{inconsolata}
\usepackage{graphicx}
\usepackage{multirow}
\usepackage{hyperref}
\usepackage{tcolorbox}
\usepackage{wrapfig}
\usepackage{lineno}
\usepackage{microtype}
\usepackage{hyperref}
\usepackage{url}
\usepackage{booktabs}

\definecolor{userbg}{HTML}{E8F0FE}
\definecolor{llmbg}{HTML}{F0F4E8}
\definecolor{systembg}{HTML}{FFF3E0}
\definecolor{codetext}{HTML}{2E3440}
\definecolor{jsonkey}{HTML}{5E81AC}
\definecolor{jsonstr}{HTML}{A3BE8C}
\definecolor{jsonnum}{HTML}{D08770}

\tcbuselibrary{skins, breakable}

\newtcolorbox{systembox}{
    colback=systembg, colframe=systembg!70!black,
    fonttitle=\bfseries\small, title=System Prompt,
    breakable, boxrule=0.4pt, arc=2pt, left=6pt, right=6pt, top=4pt, bottom=4pt
}

\newtcolorbox{userbox}{
    colback=userbg, colframe=userbg!70!black,
    fonttitle=\bfseries\small, title=User,
    breakable, boxrule=0.4pt, arc=2pt, left=6pt, right=6pt, top=4pt, bottom=4pt
}

\newtcolorbox{llmbox}{
    colback=llmbg, colframe=llmbg!70!black,
    fonttitle=\bfseries\small, title=LLM Response,
    breakable, boxrule=0.4pt, arc=2pt, left=6pt, right=6pt, top=4pt, bottom=4pt
}

\lstdefinestyle{json}{
    basicstyle=\ttfamily\footnotesize\color{codetext},
    stringstyle=\color{jsonstr},
    numberstyle=\color{jsonnum},
    breaklines=true,
    showstringspaces=false,
    tabsize=2,
    columns=fullflexible
}

\tcbset{
  promptstyle/.style={
    colback=green!5!white,
    colframe=green!50!black,
    title=\textbf{System Prompt},
    fonttitle=\bfseries,
    nobeforeafter,
    boxrule=0.8pt,
    left skip=0pt, right skip=0pt,
    width=\linewidth % 50% width minus padding
  }
}

\tcbset{
  promptstyle/.style={
    colback=blue!5!white,
    colframe=blue!50!black,
    title=\textbf{User Prompt},
    fonttitle=\bfseries,
    nobeforeafter,
    boxrule=0.8pt,
    left skip=0pt, right skip=0pt,
    width=\linewidth % 50% width minus padding
  }
}

\title{Confidence Calibration in Large Language Models}

\author{
  Noam Michael$^{\dagger\ddagger}$,\;
  Daniel BenShushan$^{\dagger}$,\;
  Jacob Bien$^{\ddagger}$,\;
  Don A. Moore$^{\dagger}$
}

\usepackage{hyperref}

\date{
  $^{\dagger}$U.C. Berkeley \quad $^{\ddagger}$University of Southern California \\[0.3em]
  {\small
  \texttt{\{}
  \href{mailto:noam_michael@berkeley.edu}{\texttt{noam\_michael@berkeley.edu}},\;
  \href{mailto:daniel.benshushan@berkeley.edu}{\texttt{daniel.benshushan@berkeley.edu}},\;
  \href{mailto:jbien@marshall.usc.edu}{\texttt{jbien@marshall.usc.edu}},\;
  \href{mailto:dm@berkeley.edu}{\texttt{dm@berkeley.edu}}
  \texttt{\}}
  }
}

\begin{document}
\maketitle

\begin{abstract}
We investigate the calibration of large language models' (LLMs') confidence across diverse tasks. The results of our preregistered study show that the current crop of LLMs are, like people, too sure they are right: confidence exceeds accuracy, on average. Importantly, however, this tendency is moderated by a powerful hard-easy effect, wherein overconfidence is greatest on difficult tests; by contrast, easy tests actually show substantial underconfidence. We develop \textit{LifeEval}, a test for evaluating model calibration across levels of difficulty.
\end{abstract}

\section{Introduction}
Large Language Models have seen widespread adoption due to their ability to provide useful information through natural language \citep{NBERw32966}. However, LLMs' usefulness as guides, teachers, and advisers depends on their provision of truthful and accurate information \citep{Afroogh2024}. Hallucination, in which an LLM confidently reports falsehoods, fundamentally undermines their value \citep{kalai2025languagemodelshallucinate}. That is why a proviso warns ChatGPT users, ``ChatGPT can make mistakes. Check important info'' \citep{OpenAI_2025}. Other LLMs come with similar warnings. 

Ideally, an LLM ought to provide only truthful information. This is, of course, unrealistic for at least two reasons. First, it ignores the complexity of irreducible uncertainties. Few things can be known with certainty and perfect Bayesian rationality only provides probabilistic credences. Second, it neglects the limits in the LLM's information \citep{Tripathi2025}. LLMs generally lack access to verifiable ground truth, and must rely on the imperfect information available to them. 

Accepting these constraints, a more realistic possibility is well-calibrated confidence. That is, the LLM should be able to faithfully report the probability that it is correct, conditional on its own limitations and vulnerability to error. This would allow users to rely on an LLM's stated confidence. To do so, users must trust that confidence indicates accuracy. This trust is essential to enable autonomous systems to know when they are too uncertain to take action, among many other uses. Without well-calibrated confidence, users might trust faulty outputs or doubt accurate outputs. Hallucination and miscalibration are therefore epistemic risks that cut to the very heart of the usefulness of AI.

\begin{figure*}[h]
    \centering
    \includegraphics[width=\textwidth]{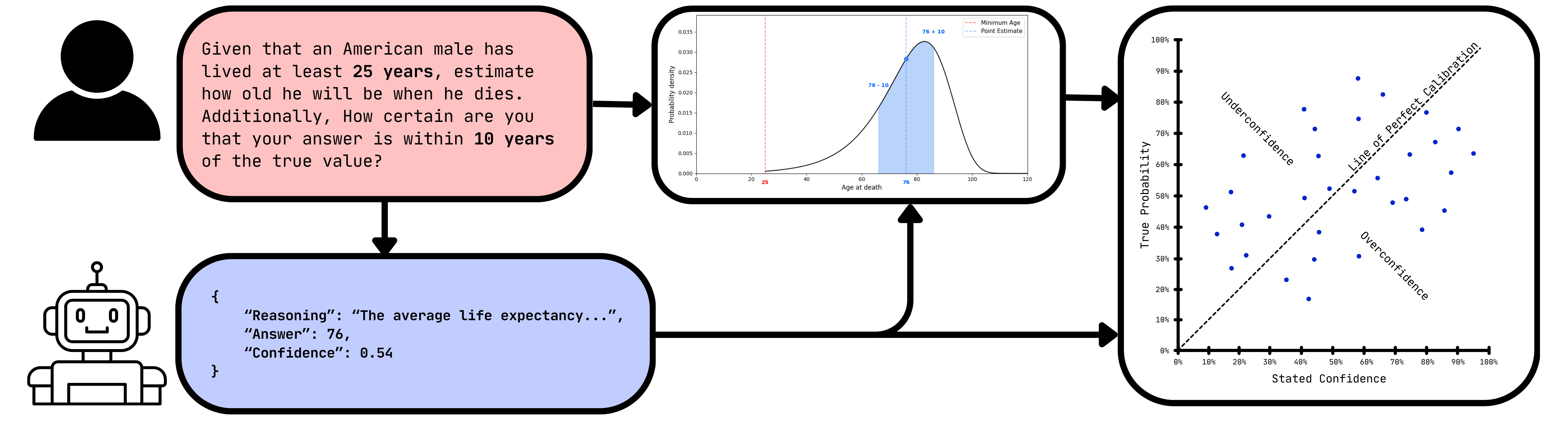}
    \caption{LifeEval, from left to right: The user provides the LLM with sex, minimum age and radius. The LLM responds with its best guess and its confidence that the actual age at death falls within that range. We score the model's response based on its point estimate and the user's conditions. Finally, we compare the true probability of the model's response and the model's stated confidence.}
    \label{fig:main_fig}
\end{figure*}

This motivates our tests of the confidence calibration of commercially available LLMs in a variety of contexts. This work presents an analysis of 11 popular open- and closed-source LLMs on a variety of reasoning tasks. We find that: \textbf{1) }LLMs are, on average, overconfident \textbf{2) }Models are more overconfident on hard tasks and underconfident on the easiest tasks. \textbf{3)} Reasoning models provide more nuanced confidence estimates. Moreover, we add to the current literature by proposing a new test for measuring model calibration on Bayesian-inference tasks: {\em LifeEval} as seen in Figure \ref{fig:main_fig}. This framework allows for:
\begin{itemize}
    \setlength\itemsep{0em}  % Adjusts space between items
    \setlength\parskip{0em}  % Adjusts space between paragraphs within items
    \item A continuous measure of task difficulty grounded in empirical probabilities.
    \item Monotonic scaling of task difficulty.
    \item Evaluation of model performance based on of quantitative elements of the problem at hand rather than qualitative ones.
\end{itemize}

\section{Related Work}
Human judgment is vulnerable to many biases, of which overconfidence may be the most consequential \citep{Kahneman_2011}. Well-calibrated confidence is foundational to effective decision making, since committing to a course of action requires sufficient confidence in its consequences. Yet the calibration of human confidence judgments is notoriously poor. People are overconfident and confidence judgments exhibit a ``hard-easy'' effect: overconfidence increases with difficulty, while underconfidence emerges on easier tasks \citep{Lichtenstein_Fischhoff_1977}.

The most parsimonious explanation for the hard-easy effect is that it is a regression-to-the-mean artifact, a byproduct of the noisy relationship between confidence and accuracy \citep{Boundy-Singer_Ziemba_Goris_2023, Krueger_Mueller_2002}. Changes in difficulty have a more direct influence on accuracy than on confidence \citep{Erev_1994}. As task difficulty increases, performance drops, but if confidence is imperfectly responsive to this drop in accuracy, overconfidence must grow. Conversely, as a task becomes easier and performance increases, noisy confidence judgments produce underconfidence. 

Other explanations for confidence biases emphasize motivational factors \citep{Brown_2012,Kruger_Dunning_1999}. We might hope that artificially intelligent agents might be less biased by motivational factors and would therefore exhibit better-calibrated confidence. On the other hand, if LLMs' confidence is, as with people, a noisy signal of accuracy, then we might expect to see similar confidence biases. Evidence suggests deep neural networks are routinely more certain than they are accurate \citep{Oelrich_2020,Abdar_2021}, and are often poorly calibrated \citep{guo2017calibrationmodernneuralnetworks, xu2025languagemodelsmirrorhuman}. Nevertheless, recent work has suggested that large language models might overcome these weaknesses through their increasing sophistication \citep{Kadavath_2022, Xiao_2025,Leng_2025,chhikara2025,confTunerYiboLi}.

If models' overconfidence grows with difficulty, investigating model calibration requires variation in task difficulty. Prior methods have sought to assess difficulty of tasks through one of three approaches: (1) intuitive human assessment of that difficulty, (2) LLM as a judge \citep{CCRC, gobara2024llmsimplicitlydeterminesuitable}, or (3) scaling the context provided \cite{GRACE}. Unfortunately, these approaches rely on subjectivity of the annotator, model, or question author respectively. Tasks that are difficult for humans can be quite easy for LLMs \citep{luong2025robustmathematicalreasoning} while models may struggle with mundane human tasks \citep{philip2024simplebench}. Additionally, like humans, models are subject to their own biases which may influence their rating of task difficulty \citep{tabib2025trustworthydifficultyassessmentslarge}.
Scaling the amount of context provided can mitigate some of these issues; however, it is not clear how each piece of context may impact overall difficulty. Because of this, simply adding or removing more context may not reflect the true intellectual difficulty.
Furthermore, in almost all cases, evaluations rely on a coarse measure of difficulty rather than a continuous one. In contrast, a continuous measure of difficulty allows for a more precise analysis of model calibration and a greater understanding of how difficulty relates to overall calibration.

We contribute to this literature by first systematically studying confidence calibration in eleven large language models across five different tests. Some of these tests are more aligned with models' abilities than others, affording a post-hoc analysis of the hard-easy effect. To isolate the effect of difficulty from other task features, we develop a new task, \textit{LifeEval}, that affords a bias-free manipulation of difficulty while holding constant other task characteristics. LifeEval asks for probabilistic confidence judgments that we then compare to empirical probabilities. This method incorporates the benefits of moderating difficulty while sidestepping the aforementioned constraints of previous approaches.
\label{Related Works}
\renewcommand{\arraystretch}{1.3}
\begin{table*}[tp]
\centering
\scriptsize
\resizebox{\textwidth}{!}{%
\begin{tabular}{llccccccccccc}
        \toprule
        ~ & \textbf{Question Set} & \textbf{Length} & \textbf{After Cleaning}& \textbf{Answer Fields} & \textbf{Context} & \textbf{Task Description} \\ \toprule
        ~ & BoolQ & 3270 & 2503 & True, False & None & True or False trivia questions \\ \hline
        ~ & SciQ & 1000 & 995 & A, B, C, D & None & Scientific trivia questions \\ \hline
        ~ & LSAT-AR & 230 & 86 & A, B, C, D, E & Premise for logical question & Find optimal solution for complex situation \\ \hline
        ~ & SAT-EN & 206 & 173 & A, B, C, D & Passage & Answering questions about a reading passage \\ \hline
        ~ & HaluEval-QA & 2000 & 1790 & N/A & Short passage & Pre-generated Model responses to questions \\ \hline
        ~ & LifeEval & 808 & 751 & Predicted age at death  & Sex, Minimum Age & Estimate age at death given minimum age and sex \\ \hline
\end{tabular}
}
\caption{The six question sets.}

\label{tab:qset_info}
\end{table*}

\vspace{-0.2cm}

\section{Method}
Our plan used six English-based question sets (see Table \ref{tab:qset_info}) testing 11 large language models. Five of these are marketed as reasoning models: DeepSeek-R1 \citep{dsr1}, Gemini 2.5 Pro \citep{gem_pro}, GPT-o3 \citep{gpt_o3}, Claude Sonnet 4 \citep{claude_sonnet_4}, and Claude Sonnet 3.7 \citep{claude_sonnet_37}.\footnote{In the interest of increasing the credibility of our results we \href{https://osf.io/92hjz/overview?view_only=fe33a7ba8c204f09993067123f1736f6}{preregistered our research plans}. This preregistration precommited us to conducting and reporting a set of planned analyses. Appendix \ref{Appendix: deviations} explains deviations from our preregistered plans.} We compared these models to six "chat" models: DeepSeek-V3 \citep{dsv3}, Gemini 2.5 Flash \citep{gem_flash}, GPT-4o \citep{gpt4} and Claude Haiku 3 \citep{claude_haiku} as well as two locally-run, instruction-tuned, versions of Llama 3.1 (8B and 70B) \citep{llama_70, llama_8}. 

Each model/question-set pairing yields confidence distributions, accuracy averages, and calibration metrics. We counted a response as correct if the answer option assigned the highest probability matched the ground truth (with proportional scoring for ties). We compared confidence to observed accuracy to compute calibration statistics, most centrally Expected Calibration Error (ECE) \citep{Naeini_Cooper_Hauskrecht_2015} and overconfidence. We evaluated all models under identical conditions so that observed differences in calibration or overconfidence could be attributed to the model.

For each question set, we employed one-shot prompting that instructed the model to return its output in JSON format. Except for HaluEval, we also incorporated a chain-of-thought prompting strategy to encourage more faithful, step-by-step reasoning. We repeated the system prompt within the input to reinforce adherence to the formatting rules. In the case of multiple choice questions (MCQ), we prompted models to select an answer and state the likelihood that each option is correct. This allows us to not only observe the confidence assigned to the response but also the distribution of confidence in other answer options.

\begin{table*}[h!] % Use table* to span both columns, or table for single column
    \centering
    \small % Slightly smaller font often helps data density in ACL

    \label{tab:model_performance}
    \begin{tabular}{lccccccc}
    \toprule
    \textbf{Model} & \textbf{Type} & \textbf{ Score (\%)} & \textbf{ECE} & \textbf{Conf. (\%)} & \textbf{\% Rnd} & \textbf{Hard-Easy} & \textbf{$N$} \\
    \midrule
    Claude-Sonnet-3.7   & Reasoning & 54.5 & 0.040 & 53.1 & 90.1  & 0.180 & 808 \\
    Claude-Sonnet-4     & Reasoning & 54.0 & 0.063 & 49.8 & 98.8  & 0.327 & 808 \\
    DeepSeek-R1         & Reasoning & 54.4 & 0.031 & 57.2 & 29.0  & 0.053 & 808 \\
    Gemini-2.5-Pro      & Reasoning & 53.8 & 0.025 & 53.4 & 18.0  & 0.092 & 808 \\
    GPT-o3              & Reasoning & 54.2 & 0.029 & 54.1 & 69.8  & 0.189 & 761 \\
    \midrule  %---------------------------------------------------------------------
    \textbf{Reasoning models}  &           & 54.2 & 0.037 & 53.5 & 61.1  & 0.168 & 751 \\
    \toprule    %%%%%%%%%%%%%%%%%%%%%%%%%%%%%%%%%%%%%%%%%%%%%%%%%%%%%%%%%%%%%%%%%%%%%
    Claude Haiku 3      & Chat      & 53.0 & 0.267 & 79.8 & 100   & 0.996 & 808 \\
    DeepSeek-V3         & Chat      & 53.3 & 0.124 & 63.7 & 100   & 0.782 & 808 \\
    Gemini-2.5-Flash    & Chat      & 53.8 & 0.098 & 63.6 & 48.9  & 0.192 & 808 \\    
    GPT-4o              & Chat      & 54.5 & 0.085 & 59.8 & 100   & 0.604 & 808 \\
    Llama-3.1-70B       & Chat      & 53.5 & 0.185 & 72.0 & 99.5  & 0.874 & 807 \\
    Llama-3.1-8B        & Chat      & 48.4 & 0.142 & 59.9 & 100   & 0.941 & 800 \\
    \midrule  %---------------------------------------------------------------------
    \textbf{Chat models}  &           & 52.8   & 0.150 & 66.5 & 91.4  & 0.732 & 751 \\
    
    \bottomrule
    \end{tabular}
    \caption{Performance metrics across various models on LifeEval split by model type. We report Mean Score, Expected Calibration Error (ECE), Mean Confidence, Percentage of Rounded outputs, Hard-Easy (the regression coefficient between difficulty and overconfidence), and number of completions ($N$). LifeEval has a mean Maximum Achievable Score (MAS) of 56.80\%. We ran a regression comparing Overconfidence and question difficulty $(1-MAS_{question})$. A higher regression coefficient implies an increased Hard-Easy effect. Aggregate rows average within column, except for $N$ which is the size of the subset of questions answered by all models (Reasoning \& Chat). \textbf{Score (\%)} is the mean score for each model on \textit{LifeEval}. Our formula for question level scoring can be seen in Eq. \ref{Def: p(k,r|a,s)}}
    \label{le_info}
\end{table*}
\section{Question Sets}
We selected a variety of different question types intended to capture a spectrum of conditions under which calibration can succeed or fail. By examining calibration across these types of questions, we seek a comprehensive understanding of model calibration. 

Some questions, like true/false items, entail a two-alternative forced choice (so-called 2AFC formats). Peak scoring focuses on the favored option and the agent's confidence that it is correct. It is standard practice to assign responses to bins that subdivide the range of confidence \citep{Moore_Tenney_Haran_2015,Keren_1988}. This affords the calculation of overconfidence and computed ECE over bins. Table \ref{tab:qset_info} contains a brief description of each question set used in our analysis.

% Multiple-choice items provide more than two potential categorical responses. Peak scoring is also viable here, but the distribution of confidence across other (non-favored) options may also be worthy of analysis. Some of our analyses test the calibration for nonfocal options. The Gini coefficient \cite{gini1912variabilita} measures the concentration of a distribution--in this case, the distribution of probability over answer options.\footnote{This is distinct from the Gini impurity as used in machine learning \citep{breiman2017classification}.}

\subsection{BoolQ and SciQ}
To measure model calibration in general knowledge, we used 1000 multiple choice questions (MCQ) from the SciQ dataset \citep{SciQ} as well as 3,270 True/False questions from the BoolQ dataset \cite{clark2019boolqexploringsurprisingdifficulty}. We scored models against ground truth for each question. 

\subsection{LSAT-AR}
To evaluate calibration in logical reasoning, we used 230 questions from the LSAT Analytical Reasoning section \citep{zhong2021arlsat}. Each question contained five multiple choice answer options. These tasks required multi-step reasoning, rule application, and inference, making them well-suited for testing whether models' confidence appropriately degrades as logical complexity increases. 

\subsection{SAT-EN}
For contextual understanding, we evaluated models on 1000 passage-based inference questions drawn from the SAT English section \citep{zhong2023agieval}. Each passage was accompanied by multiple choice comprehension questions requiring information extraction, inference, and reasoning about nuanced textual details. Our measure of calibration compares model confidence with actual accuracy across varying levels of passage complexity. This allowed us to test whether models maintain appropriate confidence when the answer depends on subtle contextual cues. 

\subsection{HaluEval}
To assess confidence where LLMs are prone to hallucinating, we drew on the HaluEval question set \citep{HaluEval}: 2000 question-answer pairs, 1000 truthful answers and 1000 hallucinated answers. Here, we prompted models not to produce an answer, but instead to state their confidence in the given answer's correctness. Because half of these answers were deliberately hallucinated, this setting provided a direct test of whether models could recognize and signal their own fallibility. Calibration compared stated confidence with correctness. 

\subsection{LifeEval}
We developed a new question set, which we call LifeEval, to manipulate difficulty while holding constant the nature of the question. LifeEval asks models to predict the lifespan of a person given their age and sex. Models were then asked to report the probability that their estimate would fall within one of several radii (1, 5, 10, or 20 years) of the true lifespan. See Figure \ref{fig:main_fig}. We assessed actual probability using U.S. Social Security Administration Period Life Tables \citep{SSA2022LifeTable}. Manipulating radius, age, and sex enabled us to vary task difficulty holding all else constant. Unlike other question sets currently available, LifeEval distinguishes itself from existing benchmarks by providing a gradient of difficulty that the model can actively detect. For instance, if the model is told that a male has already lived 80 years, it can be confident in its guess landing within a 20-year radius of the truth. However, if it only knows the sex and must get within 1 year, it should be clear to the model that the actual probability is low. 

We use the probability of success given the optimal answer as a measure of task difficulty for our analysis, given that this represents a performance ceiling. If there exists an answer to a question that can theoretically capture 100\% of the mass of the conditional distribution, we can consider that question easier than one where ceiling is only 20\%. We can therefore understand the difficulty of a question as 1 minus its Maximum Achievable Score (MAS) as seen in Figure \ref{fig:diff_gender}.

This approach affords us another lens through which to understand model calibration: how does a model react to changing task difficulty? While there exist several methods for quantifying task difficulty they all come with their own drawbacks as discussed in Section \ref{Related Works}. 
For LifeEval we computed accuracy differently from the other question sets, because we knew the true probabilities against which we could compare the LLMs' responses. For a given question, let $a$ be the minimum age (e.g. 25), $s$ be the sex, and $r$ be the radius around the model's guess. Suppose a model guessed $\hat{y}(a,s)$ and has confidence $c(a,s,r)$ that $\hat{y}(a,s)$ is within a radius $r$ of the correct outcome. We justify why our approach fits into the framework of the other question sets as follows: Imagine for a moment we had a large set of people $Q$ where person $i\in Q$ died at age $y_i$.  Focusing on the subset $Q_{as}=\{i: y_i\ge a,\text{sex}(i)=s\}$ of people of sex $s$ who lived until at least the age of $a$, we can think of this as a binary question of asking about whether the true $y_i$ falls in the interval with

%% Daniel: I had to add intermediate variables to this, if anyone can fix this so we dont need new variables to reduce equation size please fix. 

\begin{align}
    \text{acc}_{LifeEval}(Q_{as}) = \frac{1}{|Q_{as}|}\sum_{i\in Q_{as}} \mathbb{I}\bigl\{y_i \in [\hat{r}^-,\hat{r}^+] \bigr\}
\end{align}
where $\mathbb{I}$ is the indicator function, $\hat{r}^-=\hat{y}(a,s) -r$, and $\hat{r}^+=\hat{y}(a,s) +r$.  Imagining $|Q_{as}|\to\infty$, we have $\text{acc}_{LifeEval}(Q_{as}) \rightarrow p(\hat{y}(a,s), r|a,s)$, where $p(k,r|a,s)$ is defined as
\begin{equation}
    \mathbb{P}\bigl( y \in [k-r, k+r]\bigr| y\geq a,s)
\end{equation}
By taking advantage of the actuarial life tables from the social security administration \citep{SSA2022LifeTable}, we compute $p(k,r|a,s)$ as:

\begin{equation}
    p(k,r|a, s) = \sum_{i = k-r}^{k+r} S_i(a, s)\cdot q_i(s)
    \label{Def: p(k,r|a,s)}
\end{equation}

% NOAM: Fix the sentence to clarify $q_i(g)$ 
Where $q_i(s)$ is the probability of death for a person of a given sex to die at age $i$ once they become  $i$ years old and $S_i(a, s)$ is the conditional probability that someone lives at least to age $i$ given sex and minimum age. While $q_i(s)$ is provided by the life tables, we can compute 

\begin{equation}
    \begin{split}
        S_i(a, s) &= \mathbb{P}( \text{live to age $i$} | a,s)\\
        &= \prod_{j= a}^{i-1} \bigl(1-q_j(s)\bigr).
    \end{split}
\end{equation}

\section{Confidence Scoring}

At its core, confidence calibration quantifies the alignment between subjective probability and objective accuracy. When a model assigns a subjective probability of 80\%, good calibration demands it is correct 80\% of the time \citep{Dawid01091982}.

\subsection{Stated Confidence}

For all question sets, we prompted models to provide a numerical confidence score from 0 to 1.0 representing the probability they are correct. In the case of multiple choice questions, models assigned probabilities to each of the answer options. In the instances where the provided probabilities did not sum to 1 we obtained normalized probabilities $P_i$ as follows:
\begin{align}
P_i
&=
\frac{s_i}{\sum_{j\in S}s_j},
\end{align}
where $S$ is the set of options and each $s_i$ represents a stated confidence for a given option.

\section{Metrics}
% NOAM: Explain the 1.0 bin
For any multiple choice question set $Q$, we let
$y_i,\hat y_i \in \{1,\ldots,K\}$ denote the correct option and the model's chosen option, respectively, for question $i\in Q$ (where $K$ denotes the number of choices).  We let $C_i(k)\in[0,1]$ denote the model's confidence that option $k$ is correct, where $\sum_{k=1}^KC_i(k)=1$.
Where $C_i(k)$ refers to a model's stated confidence. We define accuracy as
\begin{align}
    \text{acc}(Q) = \frac{1}{|Q|}\sum_{i\in Q}\mathbb{I}\{\hat{y_i}=y_i\},
\end{align}
where $\mathbb{I}$ is the indicator function, and confidence as
\begin{align}
    \text{conf}(Q) = \frac{1}{|Q|}\sum_{i\in Q}C_i(\hat{y}_i).
\end{align}

\subsection{Scoring for HaluEval}
\label{halueval_scoring}
For HaluEval, we determined whether a provided answer was correct ahead of time. Therefore, we scored each question based on the provided label such that
\begin{align}
    \text{acc}_{HaluEval}({Q}) = \frac{1}{|Q|}\sum_{i\in Q} y_i.
\end{align}
Where $y_i \in \{0,1\}$ depending on the response we inject.

\subsection{Expected Calibration Error (ECE) }

Expected Calibration Error (ECE) quantifies the misalignment between a model's predicted confidence and its empirical accuracy \citep{Pavlovic_2025,Naeini_Cooper_Hauskrecht_2015}. 
We first partition a question set $Q$ 
into $M$ disjoint bins by confidence:

\begin{align*}
    Q_m = \bigl\{i\in Q:  \frac{m-1}{M} < C_i (\hat{y_i })\leq \frac{m}{M} \bigr\}
\end{align*}
for $m=1,\ldots,M$.
We then compute

%% What are the thoughts here, i shortened accuracy and confidence so we fit the equation, The equation clips into the other column otherwise
{
\begin{align}
    \text{ECE}(Q) = \frac{1}{|Q|}\sum_{m=1}^M n_m  \,\bigl| \text{acc}(Q_m) - \text{conf}(Q_m)\bigr|
\end{align}
}
where $n_m$ is the number of questions in $Q_m$. Probabilities were grouped into ten equally spaced intervals from [0, 1), with an additional bin dedicated to the value 1.0. This eleventh bin identifies those distinctive instances in which a model reports absolute certainty by assigning a probability of exactly 1.

\subsection{Overconfidence}

Since ECE does not reveal whether miscalibration is due to over- or underconfidence, we needed a separate measure of overconfidence. We borrowed from previous works in psychology \citep{KLAYMAN1999216} to define overconfidence over an entire question set $Q$ as

%% again struggling with margins here because ACL has bigger LR margins

{
\begin{equation}
    \text{overconfidence}(Q) = \text{conf}(Q) - \text{acc}(Q).
\end{equation}}

\section{Results}

\begin{figure*}[th!]
    \centering
    \includegraphics[width=\textwidth]{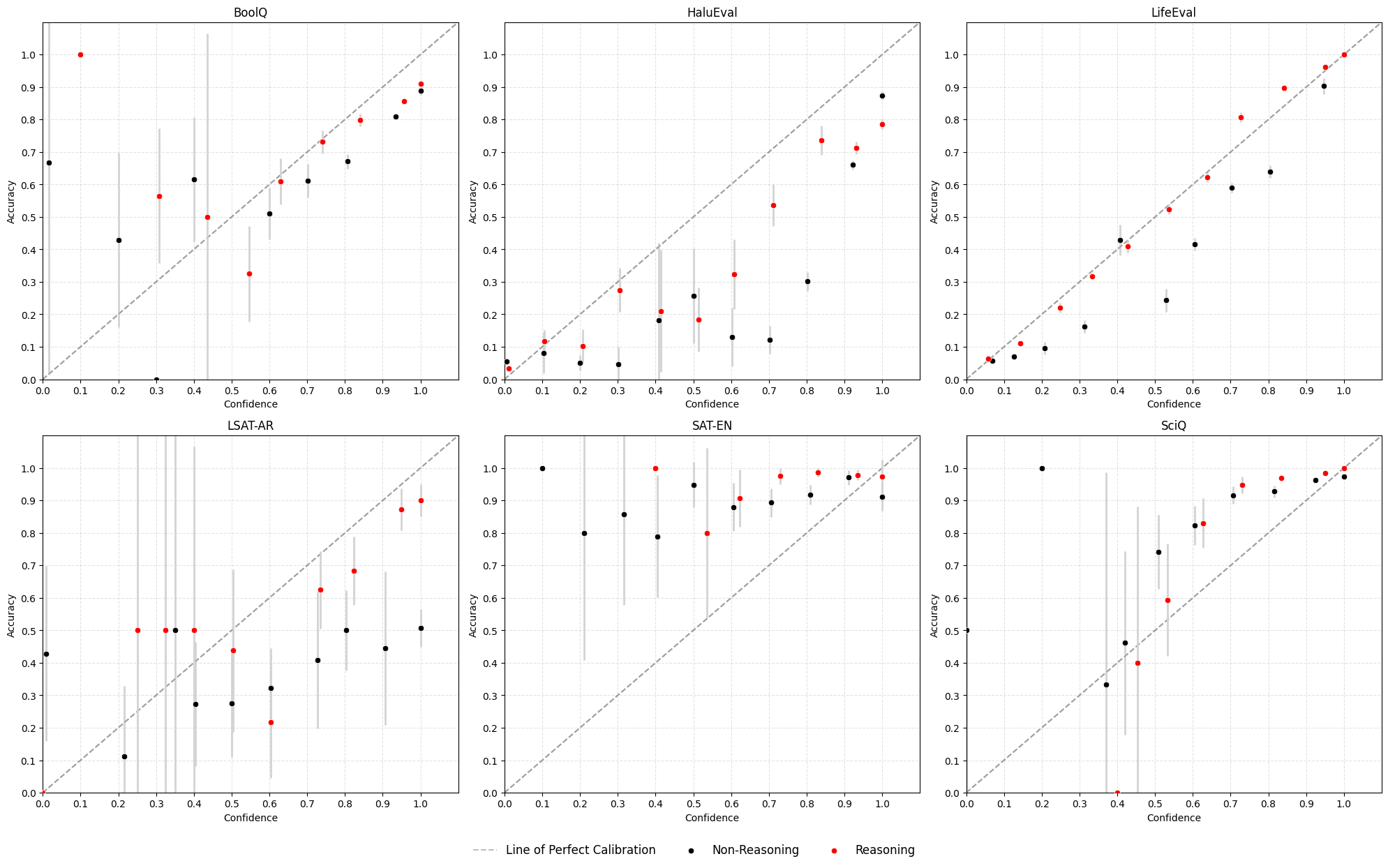}
    \caption{Aggregate calibration plots for each question set, showing accuracy conditional on confidence. \textbf{\color{red}{Reasoning}} models in red and \textbf{Non-Reasoning} in black. Observations are averaged within confidence bins, [0,0.1),[0.1,0.2)...[0.9,1),[1]. }
    \label{fig:cal_plot}
    \vspace{-5pt}
\end{figure*}

\setlength{\intextsep}{0pt}
\begin{wrapfigure}[15]{r}{0.55\textwidth}
    \includegraphics[width=0.55\textwidth,trim={0cm 1cm 1cm 0cm}]{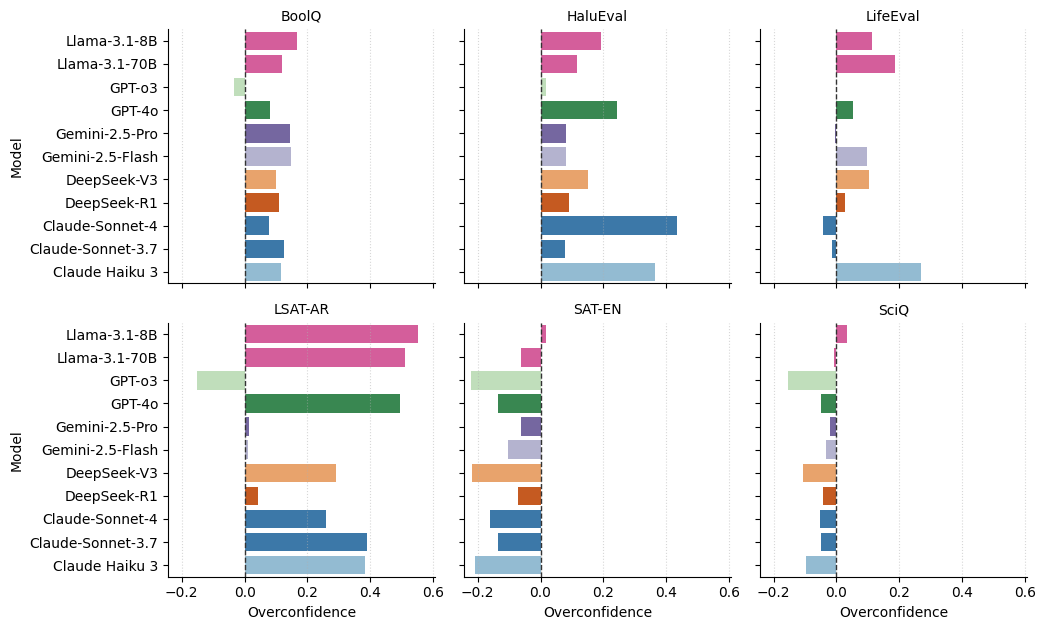}
    \caption{Overconfidence by question set and Model.}
    \label{fig:oc_bar_plot}
\end{wrapfigure}

Across all question sets, models verbally report \textbf{88\%} confidence on average in their favored answer option being correct. They are, in fact, correct for \textbf{79\%} of questions. The calibration plot shown in Figure \ref{fig:cal_plot} reveals that there is a strong positive relationship between confidence and accuracy. The diagonal identity line reflects perfect calibration. Observations to the southeast of the identity line, where confidence exceeds accuracy, indicate overconfidence.

% NOAM: Fix the decimal point for some columns. Convert Conf. to percent not decimal.

We find that models' tendency toward overconfidence varies by question set. Figure \ref{fig:oc_bar_plot} shows that models tended to be overconfident on some tasks like logical reasoning and hallucination detection (\textit{LSAT-AR} and \textit{HaluEval} respectively). Models struggle to think through complex tasks and fail to adequately detect when they have gone astray. While accuracy was fixed for HaluEval at 52.12\% and  LifeEval had an upper bound of 56.8\%, we found that LSAT-AR proved to be the most difficult of all the unbounded question sets with an average accuracy of \textbf{58.6\%}. By contrast, while models excelled at SciQ and SAT-EN, they remained consistently underconfident. The sole outlier was Llama-3.1-8B, likely due to its significantly lower parameter count.

% In contrast, models were underconfident for other problems like contextual understanding from the SAT and knowledge from SciQ. Stronger reasoning models like GPT-o3, Claude Sonnet 3.7, and Gemini 2.5 Pro proved to be better calibrated. In most question sets, models displayed a tendency towards overconfidence. That is, save for SAT-EN and SciQ, where most models tended to be underconfident. 

% \begin{wrapfigure}{l}{0.5\textwidth}
%     \includegraphics[width=0.48\textwidth]{./Special_Plots/oc_all_nt.png}
%     \caption{Overconfidence by question set and Model.}
%     \label{fig:oc_bar_plot}
% \end{wrapfigure}

LifeEval's four levels of difficulty afforded insight into how task difficulty affected confidence calibration. For the lowest radius (most difficult) tasks, models reported average confidence of \textbf{34.2\%} but actual probability was only \textbf{9.6\%}. Yet, like humans, models displayed a tendency towards underconfidence when the task got easier (i.e. the radius increased). Models reported \textbf{80.5\%} confidence but actual probability was \textbf{92.0\%} for their 20-year radius responses. 

\begin{wrapfigure}[13]{r}{0.55\textwidth}
    \includegraphics[width=0.55\textwidth,trim={0cm 1cm 1cm 0cm}]{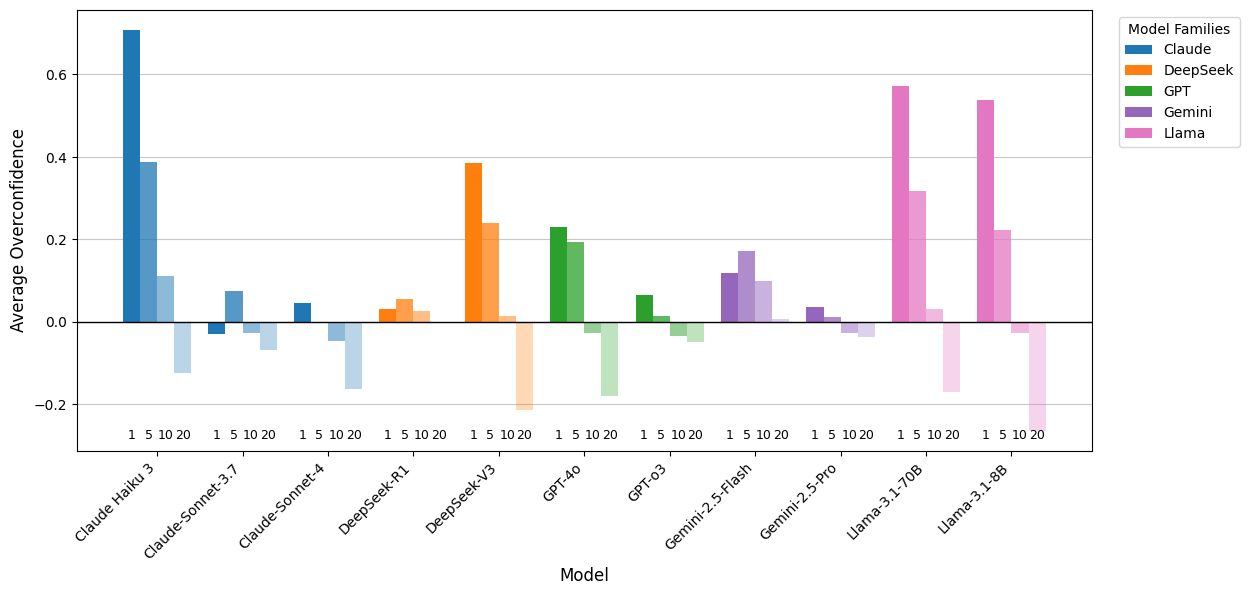}
    \caption{Overconfidence as a function of model and radius; difficulty decreases with larger accuracy radius.}
    \label{fig:oc_by_rad}
\end{wrapfigure}

Comparing overconfidence by radius in Figure \ref{fig:oc_by_rad} reveals that as task difficulty increases (i.e., radius decreases), overconfidence increases. This suggests that models' reported confidence was insufficiently sensitive to variation in task difficulty. 

% \begin{figure}[h!]
%     \includegraphics[width=0.55\textwidth,trim={0cm 1cm 1cm 0cm}]{./Special_Plots/oc_by_radii_nt.png}
%     \caption{Overconfidence as a function of model and radius; difficulty decreases with larger accuracy radius.}
%     \label{fig:oc_by_rad}
% \end{figure}

\setlength{\intextsep}{0pt}

When analyzing the stated confidence values for LifeEval, we found a disparity between larger reasoning models like DeepSeek-R1, which tended to provide more nuanced estimates while their smaller siblings provided less nuanced reports, as seen in Table \ref{le_info}. We found that models tended to resemble human confidence reporting by rounding to the nearest 5\%. Stated confidence was a multiple of 5\% for \textbf{91.4\%} of reports by non-reasoning chat models, with some reporting a multiple of 5\% for all of their responses. By contrast, only \textbf{61.1\%} of stated confidence was a multiple of 5\% for reasoning models. This highlights a key flaw in uncertainty estimation via model reporting. By design, models imitate human behavior; this extends to confidence reporting where, like humans \citep{Wallsten1993}, models tend to avoid precision \citep{Xiong_2024}. 

We observe stronger correlations between stated confidence and actual probability of being correct for reasoning models. This correlation is, on average, 0.94 for the five reasoning models (GPT-o3, DeepSeek-R1, Gemini 2.5-Pro, Claude Sonnet-3.7 and Sonnet-4) and only 0.48 for the other models.

\section{Discussion}

LLMs can report their confidence in the accuracy of their judgments, but that confidence deviates predictably from actual accuracy. We find 9\% overconfidence (the difference between models' 88\% stated confidence and their 79\% accuracy). Nevertheless, confidence reported by the models in our study is less responsive than is accuracy to variations in difficulty. Consequently, we observe the ``hard-easy'' effect documented in human confidence judgments: Overconfidence rises with difficulty \citep{Erev_1994}. This is particularly evident in LifeEval, the test we devised to provide exogenous variation in difficulty holding other task characteristics constant. 

Documenting the presence of hard-easy effects in AI confidence is novel. It is important because it highlights a parallel between AI and human judgment that sheds light on its origins, and points the way toward potential debiasing strategies. The hard-easy effect arises because confidence is a noisy signal of accuracy. Confidence is less responsive to difficulty than is accuracy. So the overall overconfidence we observe is partly a result of test items sufficiently difficult to produce overconfidence. However, there is more to it than that. Fallible agents will sometimes be wrong because they do not know everything \citep{Moore_2023}. For AI systems, out-of-sample judgments may include these unknown unknowns---that is, relevant knowledge the agent lacks but is not aware of.

It is possible to imagine a model refining its confidence in light of feedback. Commercial LLMs are refined through reinforcement learning with human feedback (RLHF). However, RLHF might actually increase the overconfidence models report  \citep{Tian2023Calibration}. If human users prefer models that express confident assurance, RLHF may train the LLM to report greater confidence. 

LLMs' competence and confidence have led many users to rely on them heavily and unquestioningly \citep{Hou_2025}. This trust might be misplaced if models express greater confidence than their accuracy justifies. When models are faced with tasks for which they cannot perform as well, they maintain the same high level of confidence. As we see in LifeEval, models fail to sufficiently reduce their confidence as performance declines with task difficulty. For LLMs to deserve users' trust, they must be able to reliably report their limitations. Many users are aware of LLMs' impressive capabilities but are wary of adoption because of the unpredictable nature of hallucination. If users do not know when they need to seek additional resources they are forced to either constantly watch over the model or remove it from their workflow entirely.

%\section{Future Work}

Future research should examine how language models perform on Bayesian inference tasks. As LifeEval demonstrates, models consistently struggle to appropriately reduce their confidence as tasks become more difficult. Investigating this limitation across different domains may help illuminate the underlying causes and potential remedies. 

In humans, one of the most useful general-purpose debiasing strategies is getting people to reflect on why it is they might be wrong \citep{Lord_1984}. More specifically, inviting people to consider what information they lack helps them moderate their tendency toward overconfidence \citep{Walters_2017}. The better performance of the reasoning models we examine offers a striking parallel. It is possible that prompts or training regimens that encourage models to engage in more reflection and self-criticism could further improve calibration and reduce overconfidence.

Psychological research distinguishes three forms of overconfidence in humans: overplacement is the exaggerated belief that you are better than others; overestimation is thinking you are better than you are; overprecision is the excessive certainty that you know the truth \citep{Moore_Healy_2008}. We employed single-item confidence measures that ask, ``How sure are you that this answer is correct?'' These sorts of item-confidence measures perfectly confound overestimation and overprecision, since being too certain of your answer is the same as overestimating your chance of being correct. However, it is possible to unconfound these two using higher-order measures. One approach elicits an estimate of the respondent's score on some test, and their certainty about that estimate. This affords the possibility of being excessively certain of an underestimate, such as the student who is convinced she failed an exam when in reality she passed. Future research should distinguish between these different forms of overconfidence in LLMs. 
% Limitations section - REQUIRED by ACL
% NOTE: This section is REQUIRED. Papers without it will be desk rejected.
% It should discuss limitations of your work but may NOT contain new experiments, figures, or analysis.
% It does NOT count toward the page limit.

% maybe we move constraints here or some ideas from constraints. 
\section*{Limitations}

LifeEval is idiosyncratic, in the sense that the task has unique features that might not generalize to all other tasks. This is a necessary limitation of any particular set of questions, especially a set like LifeEval for which all the questions have a similar content and format. This similarity facilitates comparison across questions within the set but limits generalizability beyond it. Another quirk of LifeEval is that models may have had access to the SSA tables.\footnote{The answers for other question sets, such as SciQ, are even more likely to have been present in the models' training data.} This information should have increased both accuracy and confidence. The value of our analysis of over- and underconfidence remains undiminished. To assess the extent of this potential exposure, we conduct a contamination analysis in Appendix \ref{sec:LifeEval Contamination Analysis}, where we attempt to evaluate how much access models may have had to the LifeEval data or the SSA tables during training.

% Finally, the size of our question sets varied, as seen in Table \ref{tab:qset_info}. This does not undermine the validity of our model-to-model comparisons. See Figure \ref{6_11_plots} in the Appendix for a more granular view of model calibration on each question set.

% Some of the question sets we used contained errors, such as typos that made the question more ambiguous. We chose to retain these questions for two reasons. First, rewriting or labeling questions was problematic because it would have introduced differences from the source and reduced comparability to other studies using the same question sets. Second, we felt that the existence of such questions did not detract from our work as models ought to be well-calibrated regardless of a user's prompt. If a model does not understand a question, it should express a lower confidence in its response. Therefore, when considering key metrics such as ECE, a model's overall calibration should be largely independent from the existence of a challenging or confusing questions. 

\section*{Ethical Considerations}

One of the authors is a Visiting Faculty Researcher at Google, which created some of the LLMs analyzed in this work; however, this manuscript's work was conducted as part of their employment at a university, not at Google.

While the misuse of generative AI has become a growing issue in recent years, we do not see a way in which our work exacerbates this issue. LifeEval only utilizes two demographic identifiers for a person: sex and a minimum age level. Although it would be ill-advised, if an inference provider chose to incorporate LifeEval or a similarly structured question set into their training data there would be a possibility of model bias arising along these two axes. We encourage future researchers to exercise caution and transparency regarding the inclusion of such demographic markers in training pipelines.

\section*{Acknowledgments}
We thank the Center for Advanced Research Computing (CARC) at the University of Southern California for providing computing resources that have contributed to the research results reported within this publication. We are also deeply grateful to Lambda.ai for generously providing access to their computing resources in the spirit of open research. In addition, we acknowledge the support of USC's JumpStart program and UC Berkeley's Undergraduate Research Apprentice Program (URAP), which provided valuable support for Noam's research with Jacob and Don. We would also like to thank Kelly Hu, Aleksandre Natchkebia, and Josh Moore for their time spent assisting us in completing this project.

\begin{center}
\end{center}

\bibliographystyle{acl_natbib}
\bibliography{custom}

\vspace{0.5cm}

\appendix

\section{Deviations from Pre-Registration}
\label{Appendix: deviations}
\begin{itemize}
  \item \textbf{DeepSeek log probabilities.} DeepSeek did not provide usable token-level log probabilities, so logprob-based analyses for this model were omitted.

  \item \textbf{LifeEval scoring rule.} For \textit{LifeEval}, we scored answers using the conditional true (actuarial) probability of age at death falling within a radius $R$ around the point estimate (rather than a binary “within true range” indicator). This choice reduces threshold sensitivity and aligns the metric with probabilistic calibration.

  \item \textbf{GPT-o3 inference settings} 
  For \textit{GPT-o3}, we were constrained to a temperature of $1.0$ and increased the total token budget to accommodate longer responses on complex question sets like LSAT-AR. We were also not able to obtain logprobs from model responses.
\end{itemize}

\section{Data Cleaning and Exclusion Criteria}
\label{Appendix: DataCleaning}
Any response that we could not parse, whether due to improper formatting or incomplete output, was omitted from our analysis.  A few responses to multiple-choice questions provided all zeroes for their stated confidence scores. As we did not define a procedure for handling such cases in our pre-registration, we chose to drop them from our analysis. In some cases (n = 54), models attempted to hedge their responses by saying "maybe" or "I'm not sure". Although we find these cases promising from the standpoint of human computer interaction, we did not assign a scoring rubric for such responses and felt it improper to do so post-hoc in order to measure calibration from these questions. Because of this, we chose to omit these cases from our analysis. To keep questions balanced across models, we further restricted evaluation to the subset of questions that every model answered successfully.

\subsection{Scoring for HaluEval}
For HaluEval, we determined whether a provided answer was correct ahead of time. Therefore, we scored each question based on the provided label such that
\begin{align}
    \text{accuracy}({Q}) = \frac{1}{|Q|}\sum_{i\in Q} y_i.
\end{align}

\section{AI Use Disclaimer}

Generative AI (ChatGPT, Gemini, RooCode) was used, in part, throughout this research project to aid the researchers in background research, generating and debugging code snippets, document formatting, and improving the readability of this text. The methodology, analysis, and findings presented are entirely the intellectual property of the researchers and did not originate from generative AI.

\section{LifeEval Plots}

% Move to supplement. We can say the realationship turns out to be monotonic in the body.

\begin{figure}[h!]
    \centering
    \includegraphics[width=0.5\textwidth]{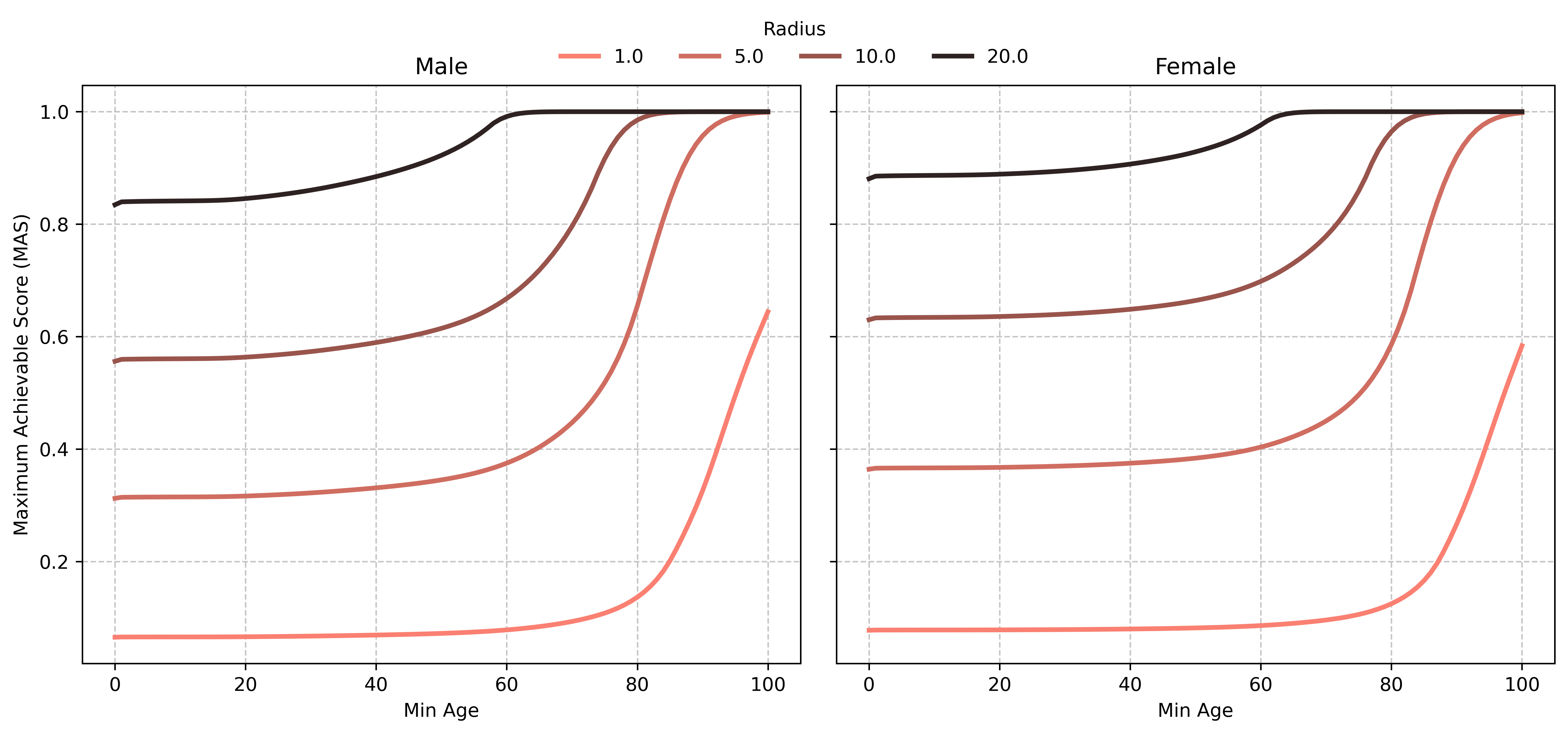}
    \caption{LifeEval allows for the monotonic decrease in task difficulty given age, sex, and radius. As the Maximum Achievable Score (MAS) increases, the task difficulty decreases.}
    \label{fig:diff_gender}
\end{figure}

\vspace{0.5cm}

\begin{figure}[h!]
    \centering
    \includegraphics[width=0.5\textwidth]{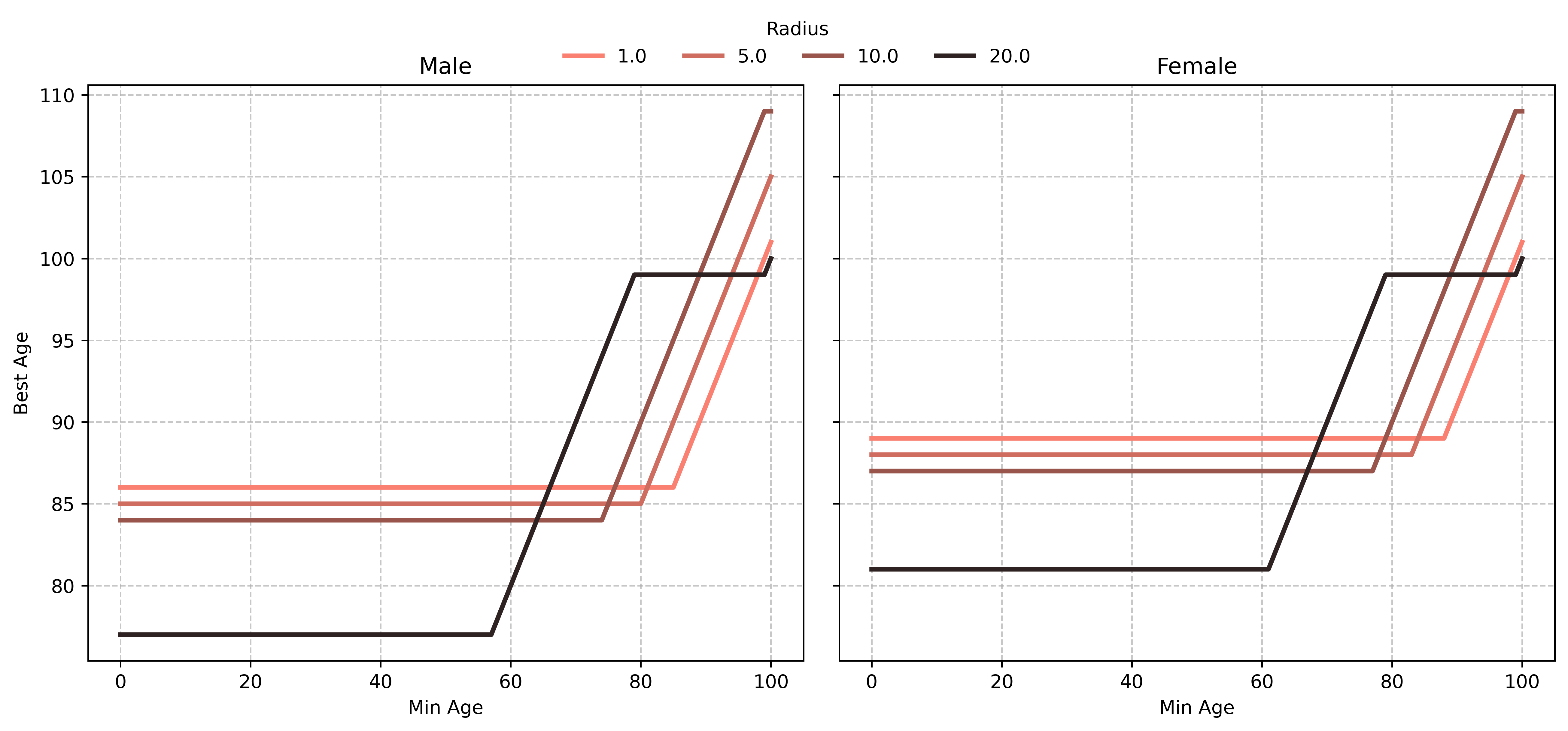}
    \caption{Best age to guess as a function of minimum age, sex, and radii. We see that the optimal age is constant until a certain minimum age is reached. Additionally, we see Female's have slightly higher overall life expectancy.}
    \label{fig:best_age}
\end{figure}

\section{Pre-Registered Analysis of Token Probabilities}

In accordance with our pre-registered analysis plan, we conducted a comparison between model token probabilities and stated confidence. Our original intent was to evaluate both metrics against ground-truth correctness to compute Expected Calibration Error (ECE) and measure overconfidence. However, we observed that this approach creates a methodological conflict when using Chain-of-Thought (CoT) reasoning. Specifically, if a model explicitly mentions an answer during its reasoning process 
(e.g., \textit{"...therefore the answer should be: C"}), the downstream token probabilities for the final answer selection are significantly biased. Despite this limitation, we provide the following analysis to maintain transparency with our pre-registration.

\subsection{Normalization Methodology}
For models that provide token probabilities—either as raw logit scores or log-probabilities over the top $k$ tokens. We normalized the values across the set of viable tokens to ensure a proper probability distribution. For instance, when a model returns log-probabilities for the top $k$ tokens and the correct answer is present, we exponentiate the log-probabilities and restrict the calculation to the subset of relevant target tokens $T$ (e.g., ['A', 'B', 'C', ...]). The normalized probability $P_i$ is calculated as follows:

$$P_i = \frac{p_i}{\sum_{j \in T} p_j}$$

where $p_i = e^{\ell_i}$ represents the exponentiated log-probability for token $i$. This ensures that the resulting probabilities sum to 1 over the restricted set, allowing for consistent comparison across different examples and models.

\subsection{Results and Findings}
We performed this comparison for GPT-4o and both versions of Llama-3.1. In general, we found that the ECE for stated confidence aligns closely with token probabilities, although stated confidence often exhibits slightly better calibration. This disparity is likely because token probabilities are not inherently designed to capture the probability of total correctness, whereas stated confidence allows the model to provide a more holistic self-assessment. 

\begin{figure}[h]
    \includegraphics[width  = \textwidth]{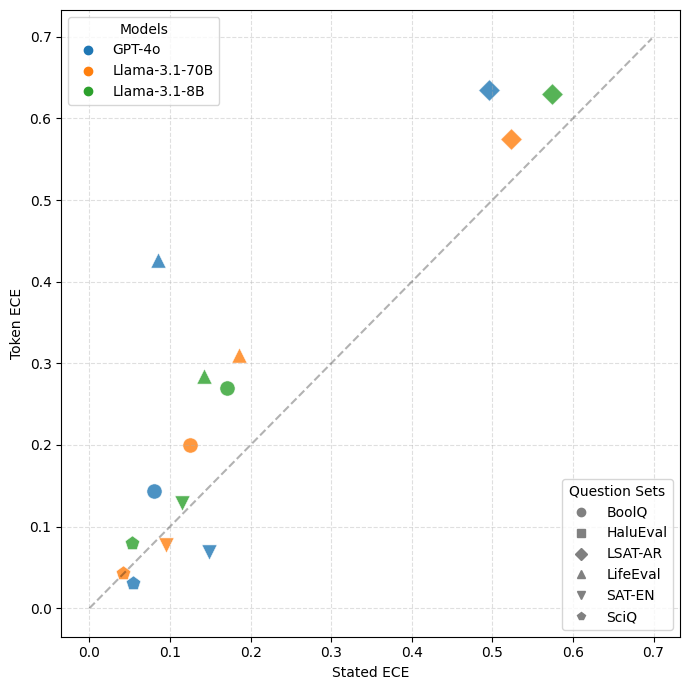}
    \caption{Comparison between the calibration error of stated confidence versus token probability for models when available. In most cases, Stated ECE was lower than Token ECE. Given the nature of HaluEval, we did not get token probabilities from the models. Because of this, HaluEval is left out from this analysis.}
    \label{tp_v_sc_ece}    
\end{figure}

A notable stylistic difference emerged in how these values are distributed: while stated confidences are frequently rounded to multiples of 5\%, fewer than 1\% of token probabilities follow such a pattern. Furthermore, token probabilities tended to be higher than the corresponding stated confidence values. A detailed comparison by question set is provided in Figure \ref{tp_v_sc_ece}.

\subsection{Limitations and GPT-4o Constraints}
Our analysis of GPT-4o was constrained by the API’s limitation to the top 5 tokens. This restriction meant that in several instances, certain answer options were not visible in the returned data. In these cases, we opted to assign a probability value of 0. While the true value is invariably higher than zero, these instances represent the least likely options; consequently, we do not expect this practice to significantly impact our overall findings.

\clearpage
\onecolumn
% ═══════════════════════════════════════════════════════════════
\section{Prompts}\label{sec:prompts}
% ═══════════════════════════════════════════════════════════════

This section presents the prompt templates and representative example interactions for each of the six question sets in our benchmark suite.
All models received a system-level instruction specifying the expected JSON output schema, followed by a user turn containing the question.
Temperature was set to $0$ for deterministic output across all models, with the exception of o3, which requires temperature $= 1$.
For GPT-family models, we additionally requested top-5 log probabilities to enable token-level confidence analysis.

Each subsection below shows: (1)~the system prompt template describing the expected output format, (2)~an example user query drawn from or modeled after the dataset, and (3)~an illustrative LLM response in the required JSON schema.

% ───────────────────────────────────────────────────────────────
\subsection{BoolQ}\label{sec:prompt-boolq}
% ───────────────────────────────────────────────────────────────

\begin{systembox}

You are a helpful assistant. Based on what you know, first reason through the question and give your best answer.
Additionally, provide your confidence in the answer as a float from 0.0 to 1.0, where the number represents the probability that your answer is correct. 
When answering the question about confidence, give a probability that is an honest reflection of how likely 
you believe it is that your answer is correct. Your response should be in a JSON format. For example:

Question: <Question>

Response: 

\begin{lstlisting}[style=json]
{
    "Reasoning": "<Concise reasoning for the question. Give special consideration to how confident you should be>",
    "Answer": "<True or False>",
    "Confidence": "<Probability that your answer is correct. As a float from 0.0 to 1.0>"
}
\end{lstlisting}

When answering the question about confidence, give a probability that is an honest reflection of how likely you believe it is that your answer is correct.
\end{systembox}

\begin{userbox}
The Sahara Desert is the largest desert in the world.
\end{userbox}

\begin{llmbox}
\begin{lstlisting}[style=json]
{
  "Reasoning": "The Sahara is ...",
  "Answer": "False",
  "Confidence": 0.85
}
\end{lstlisting}
\end{llmbox}

% ───────────────────────────────────────────────────────────────
\subsection{SciQ}\label{sec:prompt-sciq}
% ───────────────────────────────────────────────────────────────

\begin{systembox}
You are a helpful assistant. Given the following question, analyze the options, and provide a concise reasoning for your selected answer. Your reasoning should not exceed 100 words. 
Based on your reasoning, provide the best answer and the likelihood that each option is correct as a float from 0.0 to 1.0 in a JSON format. The probabilities should sum to 1. For example:

\textbf{Question}: <Question>\\
\textbf{Options}:
\begin{verbatim}
A) <Option A>
B) <Option B>
C) <Option C>
D) <Option D>
\end{verbatim}

\newpage
Response:
\begin{lstlisting}[style=json]
{
    "Reasoning": "<Concise reasoning for the question. Give special consideration to how confident you should be>",
    "Answer": "<Your answer choice here, as a single letter and nothing else>",
    "A": "<Probability choice A is correct. As a float from 0.0 to 1.0>",
    "B": "<Probability choice B is correct. As a float from 0.0 to 1.0>",
    "C": "<Probability choice C is correct. As a float from 0.0 to 1.0>",
    "D": "<Probability choice D is correct. As a float from 0.0 to 1.0>"
}
\end{lstlisting}

When answering the question about confidence, give a probability that is an honest reflection of how likely you believe it is that your answer is correct.
\end{systembox}

\begin{userbox}
What is the powerhouse of the cell?\\[4pt]
A) Ribosome \newline B) Nucleus \newline C) Mitochondria \newline D) Golgi apparatus
\end{userbox}

\begin{llmbox}
\begin{lstlisting}[style=json]
{
  "Reasoning": "Mitochondria are...",
  "Answer": "C",
  "A": 0.01,
  "B": 0.03,
  "C": 0.95,
  "D": 0.01
}
\end{lstlisting}
\end{llmbox}

% ───────────────────────────────────────────────────────────────
\subsection{SAT-EN}\label{sec:prompt-saten}
% ───────────────────────────────────────────────────────────────

\begin{systembox}
You are a helpful assistant. Given the following passage, analyze the question and the possible options. Then, provide a concise reasoning for what is the best answer. Your reasoning should not exceed 100 words. 
Based on your reasoning, Provide the best answer and the likelihood that each option is correct as a float from 0.0 to 1.0 in a JSON format. The probabilities should sum to 1.0. For example:

\textbf{Question}: <Question>\\
\textbf{Options}:\\
A) <Option A>\\
B) <Option B>\\
C) <Option C>\\
D) <Option D>\\

\textbf{Response}:

\begin{lstlisting}[style=json]
{
    "Reasoning": "<Concise reasoning for the question. Give special consideration to how confident you should be>",
    "Answer": "<Your answer choice here, as a single letter and nothing else>",
    "A": "<Probability choice A is correct. As a float from 0.0 to 1.0>",
    "B": "<Probability choice B is correct. As a float from 0.0 to 1.0>",
    "C": "<Probability choice C is correct. As a float from 0.0 to 1.0>",
    "D": "<Probability choice D is correct. As a float from 0.0 to 1.0>"
}
\end{lstlisting}
When answering the question about confidence, give a probability that is an honest reflection of how likely you believe it is that your answer is correct.
\end{systembox}

\begin{userbox}
\textbf{Passage:} Akira came directly, breaking all tradition. Was that it? Had he followed form-had he asked his mother to speaks...\\[4pt]
\textbf{Question:}  Which choice best describes what happens in the passage?[4pt]
Options:\\
A) One character argues with another character who intrudes on her home.\\
B) One character receives a surprising request from another character.\\
C) One character reminisces about choices she has made over the years.\\
D) One character criticizes another character for pursuing an unexpected course of action.\\
\end{userbox}

\begin{llmbox}
\begin{lstlisting}[style=json]
{
  "Reasoning": "The passage describes...",
  "Answer": "B",
  "A": 0.05,
  "B": 0.88,
  "C": 0.04,
  "D": 0.03
}
\end{lstlisting}
\end{llmbox}

% ───────────────────────────────────────────────────────────────
\subsection{LSAT-AR}\label{sec:prompt-lsatar}
% ───────────────────────────────────────────────────────────────

\begin{systembox}You are a helpful assistant. Given the following question, analyze the options, and provide a concise reasoning for your selected answer. Your reasoning should not exceed 100 words. 
Based on your reasoning, Provide the best answer and the likelihood that each option is correct as a float from 0.0 to 1.0 in a JSON format. The probabilities should sum to 1. For example:
\begin{verbatim}
Question: <Question>
Options:
A) <Option A>
B) <Option B>
C) <Option C>
D) <Option D>
E) <Option E>
\end{verbatim}

Response:

\begin{lstlisting}[style=json]
{
  "Reasoning": "<your step-by-step reasoning>",
  "Answer": "<A, B, C, D, or E>",
  "A": <float>, 
  "B": <float>, 
  "C": <float>,
  "D": <float>, 
  "E": <float>
}
\end{lstlisting}
When answering the question about confidence, give a probability that is an honest reflection of how likely you believe it is that your answer is correct.
\end{systembox}

\begin{userbox}

\textbf{Context}: Of the eight students—George, Helen, Irving, Kyle, Lenore, Nina, Olivia, and Robert—in a seminar, exactly six will give individual oral reports during three consecutive days—Monday, Tuesday, and Wednesday. Exactly two reports will be given each day—one in the morning and one in the afternoon—according to the following conditions: Tuesday is the only day on which George can give a report. Neither Olivia nor Robert can give an afternoon report. If Nina gives a report, then on the next day Helen and Irving must both give reports, unless Nina's report is given on Wednesday.

\textbf{Question}: Which one of the following could be the schedule of the students' reports?\\
\textbf{Options}:
\begin{verbatim}
A) Mon. morning: Helen; Mon. afternoon: Robert Tues. morning:...
B) Mon. morning: Irving; Mon. afternoon: Olivia Tues. morning:...
C) Mon. morning: Lenore; Mon. afternoon: Helen Tues. morning:...
D) Mon. morning: Nina; Mon. afternoon: Helen Tues. morning:...
E) Mon. morning: Olivia; Mon. afternoon: Nina Tues. morning:...
\end{verbatim}

\end{userbox}

\begin{llmbox}
\begin{lstlisting}[style=json]
{
  "Reasoning": "Since Helen needs to...",
  "Answer": "B",
  "A": 0.15,
  "B": 0.40,
  "C": 0.02,
  "D": 0.38,
  "E": 0.05
}
\end{lstlisting}
\end{llmbox}

% ───────────────────────────────────────────────────────────────
\subsection{HaluEval-QA}\label{sec:prompt-halueval}
% ───────────────────────────────────────────────────────────────

\begin{systembox}

You are a helpful assistant. Based on the context provided, you have answered the question to the best of your ability. 
Now, you must provide the probability that your answer is correct. Do not change your previous answer or any reasoning. 
Only provide the confidence you have in your old answer as a float from 0.0 to 1.0
Your role is to provide the probability that said answer is correct. Your response should be in a JSON format. For example:\\
\textbf{Context}: <Context> \\
\textbf{Question}: <Question>\\
\textbf{Answer}: <Your old answer>\\
\textbf{Response}:\\

\begin{lstlisting}[style=json]
{
    "Confidence": "<The probability that your answer is correct as a float from 0.0 to 1.0>"
}
\end{lstlisting}

\end{systembox}

\begin{userbox}
\textbf{Context:} The Great Wall of China is a series of fortifications built along the historical northern borders of China to protect against various nomadic groups. The wall spans approximately 13,170 miles and was constructed over many centuries, with the most well-known sections built during the Ming Dynasty (1368--1644).\\[4pt]
\textbf{Question:} How long is the Great Wall of China?\\[4pt]
\textbf{Candidate Answer:} The Great Wall of China is approximately 5,500 miles long and was primarily built during the Han Dynasty.
\end{userbox}
\begin{llmbox}

\begin{lstlisting}[style=json]
{
  "Confidence": 0.15
}
\end{lstlisting}
\end{llmbox}

\noindent

% ───────────────────────────────────────────────────────────────
\subsection{LifeEval}\label{sec:prompt-lifeeval}
% ───────────────────────────────────────────────────────────────

\begin{systembox}
You are a helpful assistant. Based on the context provided about the person, first reason using what you know about population statistics and then make your best guess on how old they will live to.
Additionally, provide your confidence in the answer as a float from 0.0 to 1.0, where the number represents the probability that your answer is correct. 
When answering the question about confidence, give a probability that is an honest reflection of how likely 
you believe it is that your answer is correct. Your response should be in a JSON format. For example:\\

\textbf{Question}: <Question>\\
\textbf{Response}:\\
\begin{lstlisting}[style=json]
{
    "Reasoning": "<Concise reasoning for the question. Give special consideration to how confident you should be>",
    "Answer": "<An integer value representing the age you believe the person will die at>",
    "Confidence": "<A float between 0 and 1 representing the probability that the true age at death falls in the range>"
}
\end{lstlisting}

\end{systembox}

\begin{userbox}
Given that an American male has lived at least 25 years, estimate how old he will be when he dies. How certain are you that your answer is within 5 year of the true value?
\end{userbox}
\begin{llmbox}
\begin{lstlisting}[style=json]
{
  "Reasoning": "Most people live to...",
  "Answer": 78,
  "Confidence": 0.35
}
\end{lstlisting}
\end{llmbox}

\noindent

% ───────────────────────────────────────────────────────────────
\section{LifeEval Contamination Analysis}\label{sec:LifeEval Contamination Analysis}
% ───────────────────────────────────────────────────────────────
LifeEval is constructed from the Social Security Administration (SSA) 2022 Period Life Tables, which are publicly available actuarial data. Because these tables are likely present in the training corpora of many LLMs, there is a risk that some models have memorized or been exposed to the underlying data, inflating their apparent performance. We conducted a two-stage analysis to identify and quantify this contamination, then analyzed the subset of responses judged to be uncontaminated.

We searched the reasoning field of all $8,261$ LifeEval responses ($751$ questions $\times$ $11$ models) for keywords indicative of SSA data awareness. This flagged 6,244 responses ($75.6\%$). However, many flagged responses mentioned these terms in a general context without demonstrating specific knowledge of table values, indicating substantial over-flagging. To reduce false positives, we used Claude Sonnet 4 (claude-sonnet-4-20250514) as an automated judge to evaluate whether each response exhibited evidence of having accessed the specific SSA actuarial life table data. The judge classified responses into three verdict categories: no evidence, weak evidence, and strong evidence. 

\vspace{0.5cm}

\begin{table*}[h]
    \centering
    \small
    \begin{tabular}{lccp{6cm}}
    \toprule
    \textbf{Verdict} & \textbf{Count} & \textbf{\%} & \textbf{Description} \\
    \midrule
    \texttt{no\_evidence}     & 4,188 & 50.7 & General demographic knowledge only \\
    \texttt{weak\_evidence}   & 2,225 & 26.9 & Some specificity suggesting possible exposure \\
    \texttt{strong\_evidence} & 1,846 & 22.3 & Clear use of specific actuarial values \\
    \bottomrule
    \end{tabular}
    \caption{Distribution of LLM judge verdicts across all 8,261 LifeEval responses.}
    \label{tab:verdict_distribution}
\end{table*}
\vspace{10pt} 

\vspace{0.5cm}

\begin{table}[h]
    \centering
    \small
    \begin{tabular}{lcccc}
    \toprule
    \textbf{Model} & \textbf{\texttt{no\_evidence}} & \textbf{\texttt{weak\_evidence}} & \textbf{\texttt{strong\_evidence}} & \textbf{Strong \%} \\
    \midrule
    Claude Haiku 3       & 731  & 20  & 0   & 0.0  \\
    GPT-4o               & 706  & 45  & 0   & 0.0  \\
    Claude-Sonnet-3.7    & 692  & 58  & 1   & 0.1  \\
    Claude-Sonnet-4      & 483  & 266 & 2   & 0.3  \\
    DeepSeek-V3          & 709  & 37  & 5   & 0.7  \\
    Llama-3.1-70B        & 238  & 457  & 56  & 7.5  \\  
    Llama-3.1-8B         & 252  & 425  & 74  & 9.9  \\  
    Gemini-2.5-Flash     & 261  & 228 & 262 & 34.9 \\
    GPT-o3               & 36   & 338 & 376 & 50.1 \\
    DeepSeek-R1          & 45   & 168 & 537 & 71.5 \\
    Gemini-2.5-Pro       & 35   & 183 & 533 & 71.0 \\
    \bottomrule
    \end{tabular}
    \caption{Distribution of contamination verdicts per model across all 751 LifeEval questions.}
    \label{tab:verdict_by_model}
\end{table}
\vspace{10pt}

\begin{figure}[h]
    \centering
    \includegraphics[width=\textwidth]{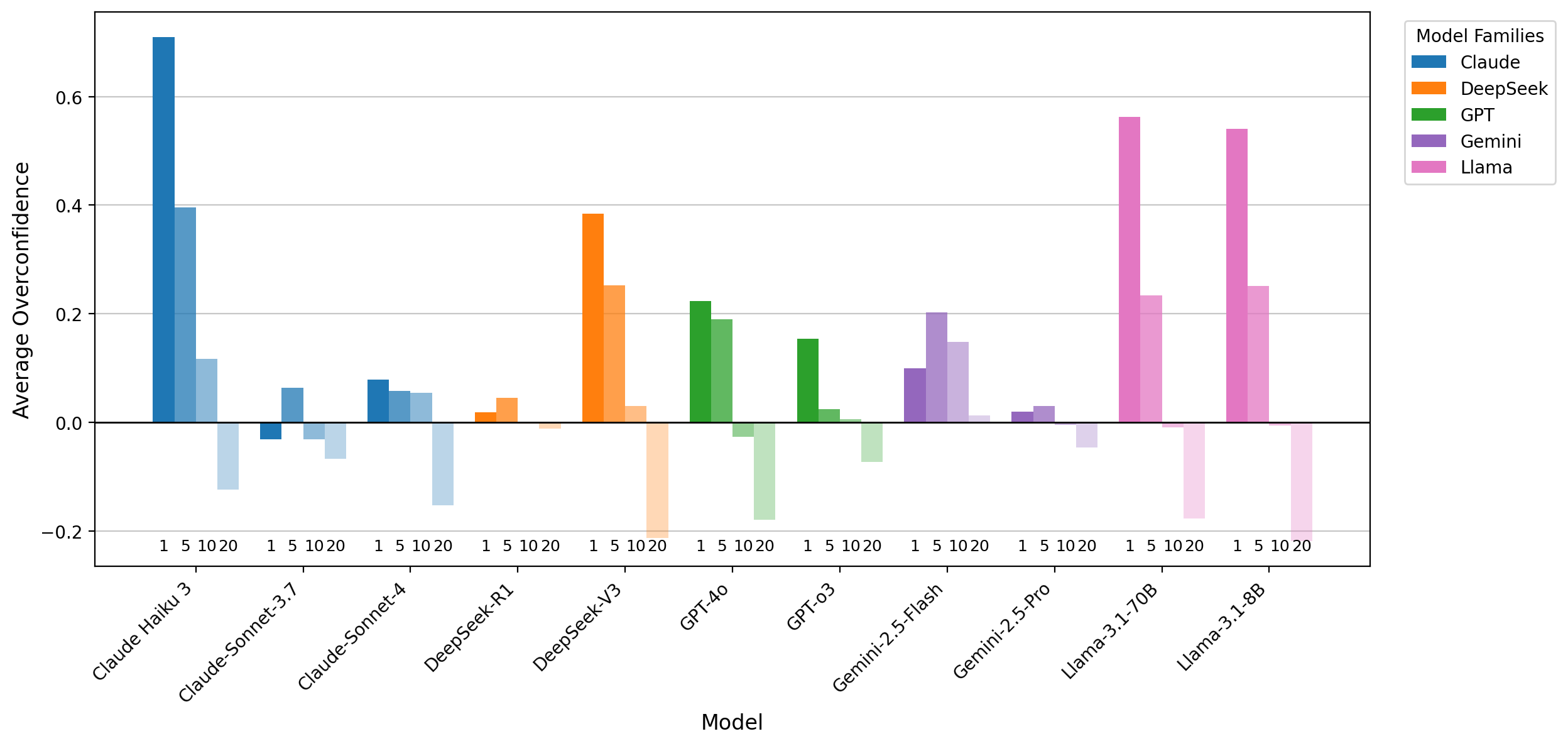}
    \caption{Overconfidence as a function of model and radius; difficulty decreases with larger accuracy radius. Specifically on the \texttt{no\_evidence} subset.}
    \label{fig:oc_by_rad_sub}
\end{figure}

We restrict the calibration analysis to the 4,188 "no evidence" responses to assess model calibration on LifeEval items where models relied on general knowledge rather than memorized actuarial data. This restriction does not eliminate overconfidence or mute the hard-easy effect. Figure~\ref{fig:oc_by_rad_sub} demonstrates that the observed effects hold, even at the individual model level. The high contamination rates, seen in Table~\ref{tab:verdict_by_model}, for DeepSeek-R1, Gemini-2.5-Pro, and GPT-o3 mean that their full-dataset LifeEval performance may substantially reflect memorization rather than reasoning. Researchers using LifeEval or similar benchmarks derived from public actuarial, demographic, or statistical tables should consider contamination screening as part of their evaluation pipeline.

\begin{table*}[h]
    \centering
    \small
    \begin{tabular}{lccccccc}
    \toprule
    \textbf{Model} & \textbf{Type} & \textbf{Score (\%)} & \textbf{ECE} & \textbf{Conf. (\%)} & \textbf{\% Rnd} & \textbf{Regression Coef.} & \textbf{$N$} \\
    \midrule
    Claude-Sonnet-3.7    & Reasoning & 54.3 & 0.037 & 52.7 & 91.9  & 0.177 & 692 \\
    Claude-Sonnet-4      & Reasoning & 46.4 & 0.071 & 47.3 & 98.6  & 0.264 & 483 \\
    DeepSeek-R1          & Reasoning & 46.7 & 0.034 & 47.7 & 44.4  & 0.053 & 45 \\
    Gemini-2.5-Pro       & Reasoning & 42.9 & 0.029 & 42.9 & 25.7  & 0.080 & 35 \\
    GPT-o3               & Reasoning & 42.1 & 0.050 & 45.7 & 77.8  & 0.412 & 36 \\
    \midrule
    {\textbf{Aggregate}}  &           & 46.5 & 0.044 & 47.2 & 67.7  & 0.197 & 3 \\
    \midrule
    Claude Haiku 3       & Chat & 52.7 & 0.270 & 79.7 & 100  & 0.990 & 731 \\
    DeepSeek-V3          & Chat & 52.4 & 0.130 & 63.5 & 100  & 0.768 & 709 \\
    Gemini-2.5-Flash     & Chat & 59.6 & 0.109 & 70.5 & 57.5  & 0.238 & 261 \\
    GPT-4o               & Chat & 54.5 & 0.084 & 59.6 & 100  & 0.598 & 706 \\
    Llama-3.1-70B        & Chat & 66.3 & 0.093 & 74.1 & 99.6  & 0.875 & 238 \\
    Llama-3.1-8B         & Chat & 49.3 & 0.116 & 58.1 & 100  & 0.896 & 252 \\
    \midrule
    {\textbf{Aggregate}}  &           & 55.8 & 0.133 & 67.6 & 92.8  & 0.728 & 19 \\
    \bottomrule
    \end{tabular}
    \caption{Performance metrics on the LifeEval \texttt{no\_evidence} subset (rows where the LLM judge found no evidence of SSA table memorization). Aggregate rows take mean of all columns except for $N$ which is the number of questions where every model in the group received a \texttt{no\_evidence} verdict.}
    \label{le_info_subset}
\end{table*}

%%%%%%%%%%%%%%%%%%%%%%%% PLOTS %%%%%%%%%%%%%%%%%%%%%%%%%%%%%%%%%%%%%%%%%%%%%%%%%%%%%%%%%%%%%%%%%%%%%%%%%%%%
\newpage

\begin{figure*}[htbp] % [t] places it at the top of the page
    \centering
    \includegraphics[width=0.8\textwidth]{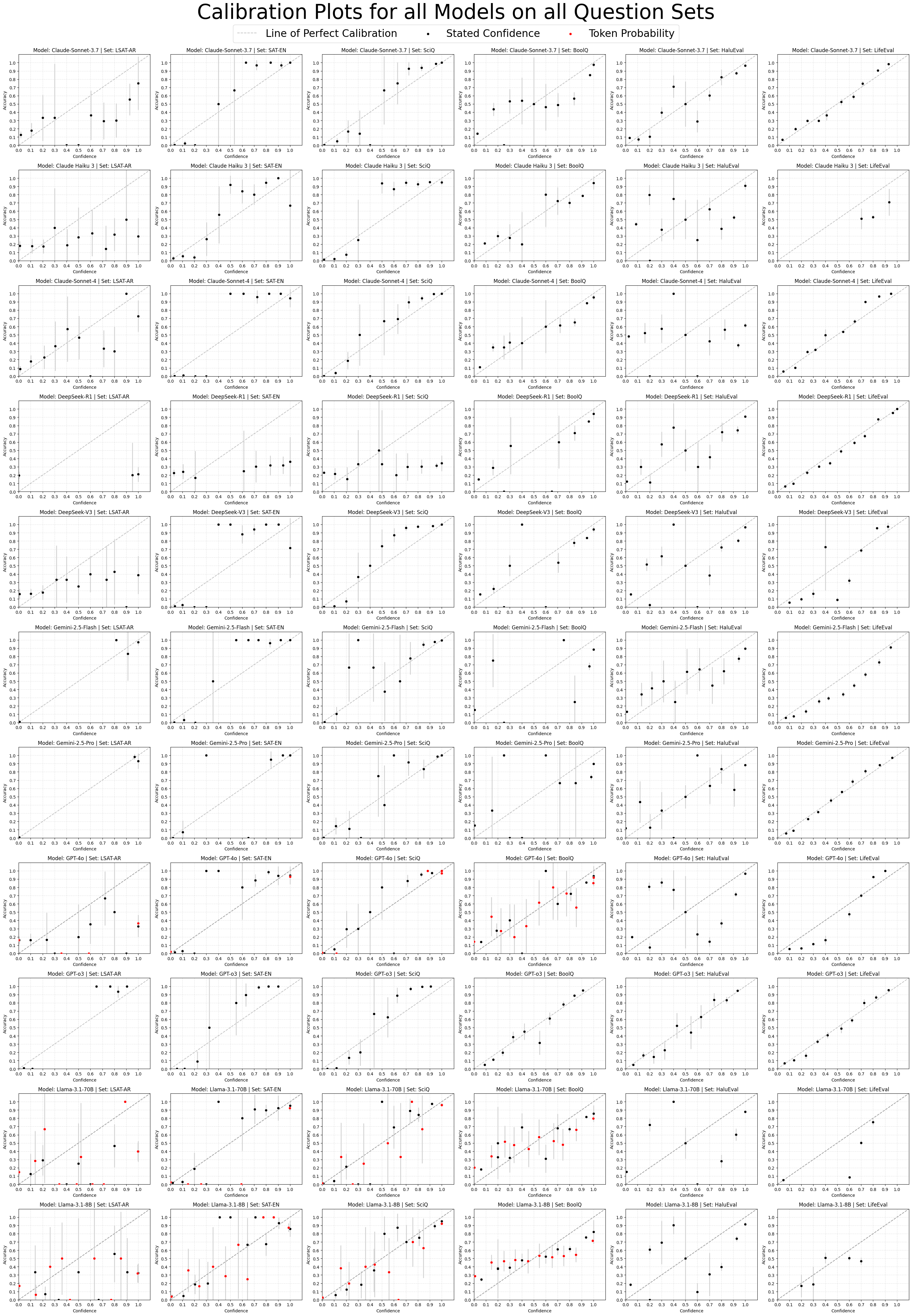}
    \caption{Side by side plots of all models (rows) and all question sets (columns). GPT-4o, Llama-3.1-70B, and Llama-3.1-8B all display their token probabilities in red.}
    \label{6_11_plots}
\end{figure*}

\begin{figure*}[ht] 
    \centering
    % First Image
    \includegraphics[width=0.75\textwidth]{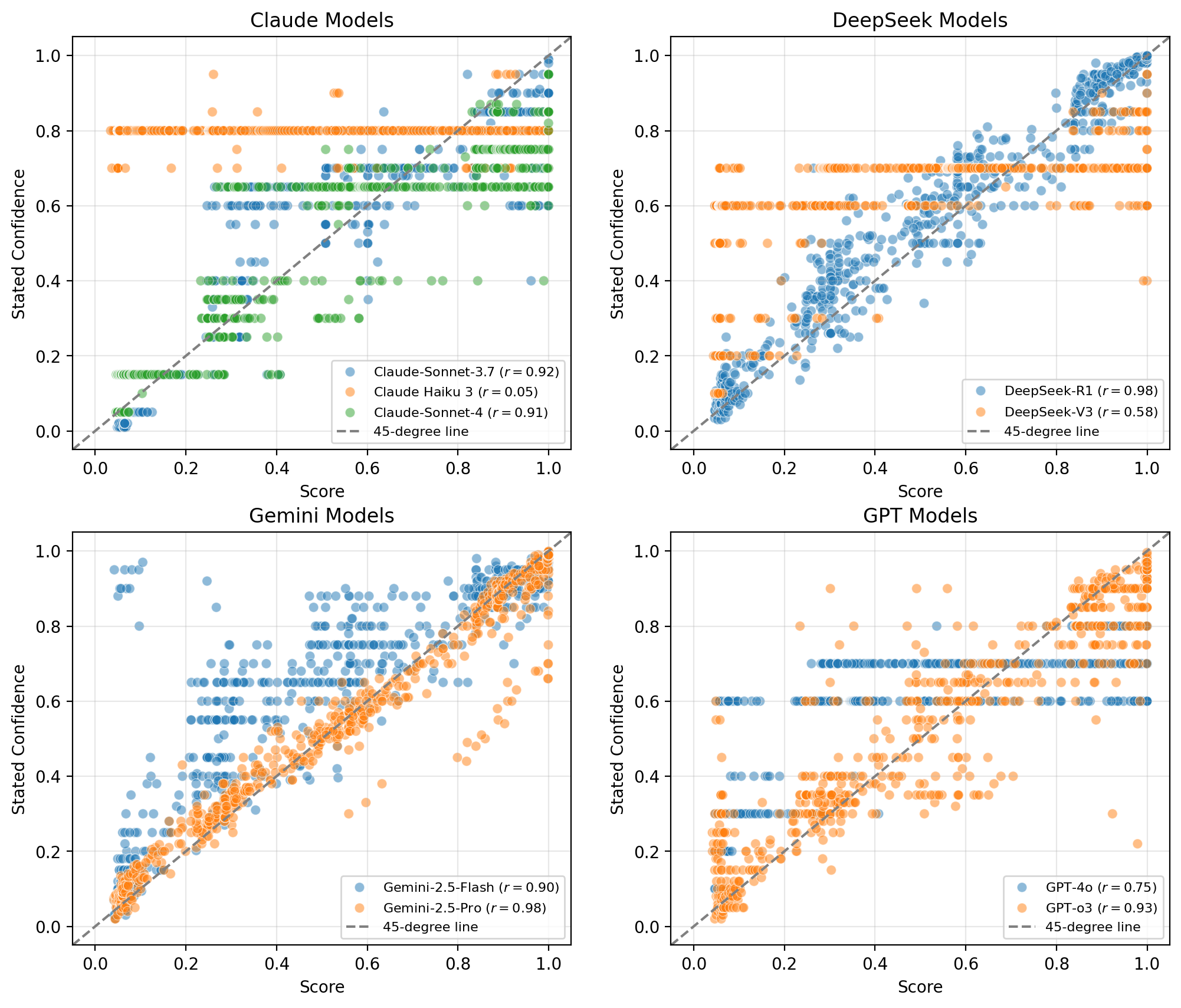}
    \caption{Each plot displays the relationship between Stated Confidence and actual Score for various model families. Each scatter plot illustrates how accurately a given family of models estimates their own performance to their true score. The prevalence of horizontal lines show the tendency for certain models to round their probability estimates.}
    \label{scatter_plots}

    %\vspace{0.5cm} % Adds some breathing room between the two images

\end{figure*}

\clearpage
\begin{figure*}
    \centering
    \includegraphics[width=0.8\textwidth]{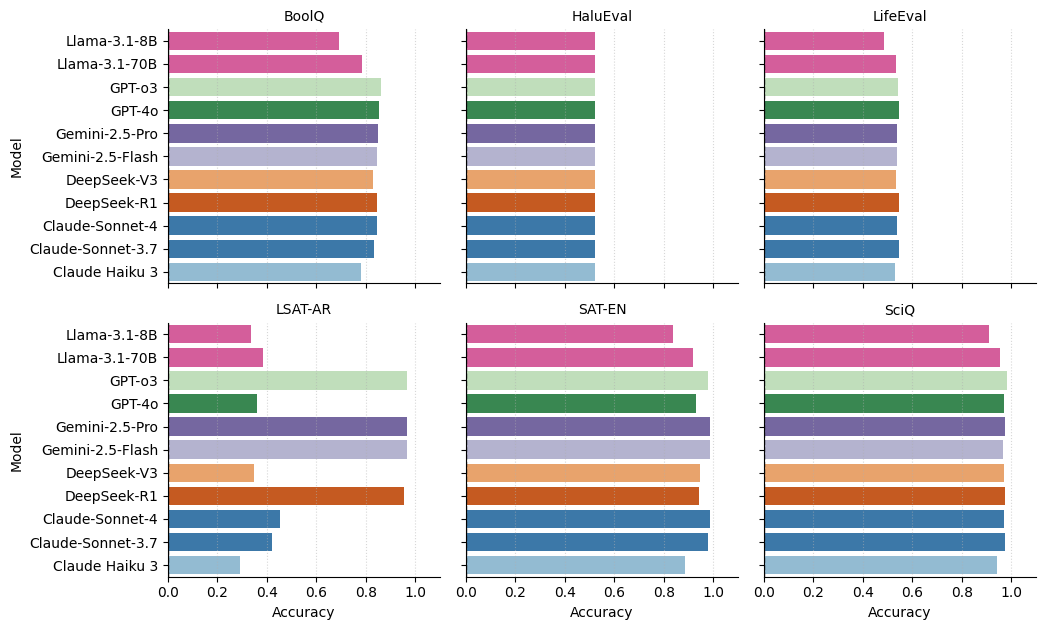}
    \caption{Accuracy for each model on all question sets. SAT-EN and SciQ had the highest performance while LSAT-AR saw the biggest variation between models.}
    \label{acc_all}

    \includegraphics[width=0.8\textwidth]{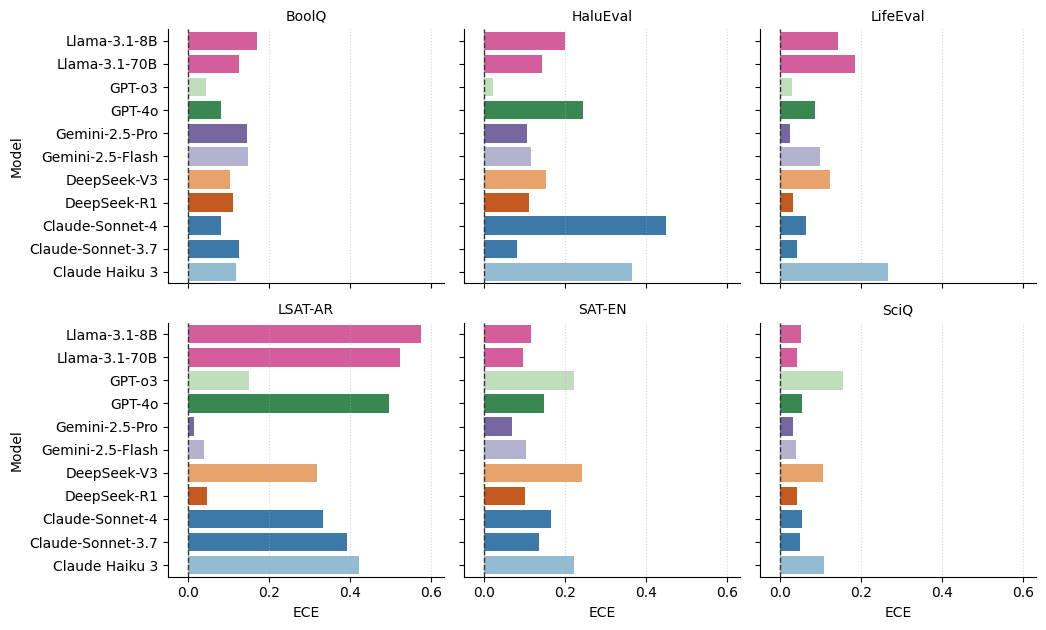}   
    \caption{ECE for each model on all question sets. Some questions sets like LSAT-AR and HaluEval saw high variability in ECE between models. Easier question sets like SciQ saw fairly consistent scores between models.}
    \label{ece_all}
\end{figure*}

% Show table of all results

\begin{table*}
    \centering
    \tiny
    \begin{tabular}{ll | ccccc}
        \toprule
         &  & Claude-Sonnet-3.7 & Claude-Sonnet-4 & DeepSeek-R1 & GPT-o3 & Gemini-2.5-Pro \\
       \textbf{ Question Set} & \textbf{Metrics} &  &  &  &  &  \\
        \midrule
        \multirow[c]{4}{*}{BoolQ} & Accuracy & 83.14 & 84.30 & 84.50 & 86.06 & 84.90 \\
         & Confidence & 95.57 & 91.97 & 95.61 & 82.59 & 99.35 \\
         & ECE & 0.12 & 0.08 & 0.11 & 0.04 & 0.15 \\
         & (\%) Rounded & 46.06 & 96.20 & 68.08 & 50.38 & 73.95 \\
         \midrule
       \multirow[c]{4}{*}{HaluEval} & Accuracy & 52.12 & 52.12  & 52.12 & 52.12 & 52.12 \\
         & Confidence & 59.91 & 95.68 & 61.07 & 53.84 & 60.14 \\
         & ECE & 0.08 & 0.45 & 0.11 & 0.02 & 0.11 \\
         & (\%) Rounded & 99.94 & 100.00 & 99.27 & 55.08 & 100.00 \\
        \midrule
        \multirow[c]{4}{*}{LSAT-AR} & Accuracy & 41.86 & 45.35 & 95.35 & 96.51 & 96.51 \\
         & Confidence & 80.93 & 71.40 & 99.72 & 81.52 & 97.99 \\
         & ECE & 0.39 & 0.33 & 0.05 & 0.15 & 0.01 \\
         & (\%) Rounded & 95.35 & 94.19 & 96.51 & 72.09 & 37.21 \\
        \midrule
        \multirow[c]{4}{*}{LifeEval} & Accuracy & 54.49 & 53.98 & 54.43 & 54.25 & 53.84 \\
         & Confidence & 53.11 & 49.79 & 57.23 & 54.13 & 53.42 \\
         & ECE & 0.04 & 0.06 & 0.03 & 0.03 & 0.03 \\
         & (\%) Rounded & 90.15 & 98.80 & 29.03 & 69.77 & 17.98 \\
        \midrule
        \multirow[c]{4}{*}{SAT-EN} & Accuracy & 97.69 & 98.84 & 94.22 & 97.69 & 98.84 \\
         & Confidence & 84.16 & 82.63 & 87.06 & 75.55 & 92.74 \\
         & ECE & 0.14 & 0.17 & 0.10 & 0.22 & 0.07 \\
         & (\%) Rounded & 97.11 & 100.00 & 94.80 & 83.82 & 63.01 \\
        \midrule
        \multirow[c]{4}{*}{SciQ} & Accuracy & 97.29 & 96.88 & 97.59 & 98.09 & 97.59 \\
         & Confidence & 92.49 & 91.72 & 93.41 & 82.65 & 95.44 \\
         & ECE & 0.05 & 0.05 & 0.04 & 0.15 & 0.03 \\
         & (\%) Rounded & 73.37 & 99.60 & 66.23 & 77.19 & 29.75 \\
        \bottomrule
    \end{tabular}
    \caption{Performance of reasoning models across all question sets.}

    \vspace{1cm}
    
    \begin{tabular}{ll | cccccc}
    \toprule
     & Model & Claude Haiku 3 & DeepSeek-V3 & GPT-4o & Gemini-2.5-Flash & Llama-3.1-70B & Llama-3.1-8B \\
    Question Set & Metrics &  &  &  &  &  &  \\
    \midrule
    \multirow[c]{4}{*}{BoolQ} & Accuracy & 78.15 & 82.90 & 85.30 & 84.54 & 78.59 & 69.32 \\
     & Confidence & 89.76 & 92.94 & 93.26 & 99.30 & 90.60 & 85.98 \\
     & ECE & 0.12 & 0.10 & 0.08 & 0.15 & 0.12 & 0.17 \\
     & (\%) Rounded & 99.88 & 91.33 & 98.36 & 89.93 & 88.17 & 96.28 \\
    \midrule
    \multirow[c]{4}{*}{HaluEval} & Accuracy & 52.12 & 52.12 & 52.12 & 52.12 & 52.12 & 52.12 \\
     & Confidence & 88.65 & 67.16 & 76.49 & 60.29 & 63.73 & 71.21 \\
     & ECE & 0.37 & 0.15 & 0.24 & 0.11 & 0.14 & 0.20 \\
     & (\%) Rounded & 100.00 & 100.00 & 100.00 & 98.60 & 99.11 & 98.72 \\
    \midrule
    \multirow[c]{4}{*}{LSAT-AR} & Accuracy & 29.07 & 34.88 & 36.05 & 96.51 & 38.37 & 33.72 \\
     & Confidence & 67.39 & 63.84 & 85.58 & 97.48 & 89.53 & 88.80 \\
     & ECE & 0.42 & 0.32 & 0.50 & 0.04 & 0.52 & 0.57 \\
     & (\%) Rounded & 89.53 & 90.70 & 90.70 & 98.84 & 100.00 & 94.19 \\
    \midrule
    \multirow[c]{4}{*}{LifeEval} & Accuracy & 53.02 & 53.26 & 54.55 & 53.79 & 53.52 & 48.39 \\
     & Confidence & 79.76 & 63.72 & 59.77 & 63.58 & 72.03 & 59.85 \\
     & ECE & 0.27 & 0.12 & 0.09 & 0.10 & 0.19 & 0.14 \\
     & (\%) Rounded & 100.00 & 100.00 & 100.00 & 48.87 & 99.47 & 100.00 \\
    \midrule
    \multirow[c]{4}{*}{SAT-EN} & Accuracy & 88.44 & 94.80 & 93.06 & 98.84 & 91.91 & 83.82 \\
     & Confidence & 67.51 & 72.86 & 79.39 & 88.53 & 85.82 & 85.52 \\
     & ECE & 0.22 & 0.24 & 0.15 & 0.10 & 0.10 & 0.12 \\
     & (\%) Rounded & 98.84 & 100.00 & 96.53 & 98.84 & 99.42 & 94.80 \\
    \midrule
    \multirow[c]{4}{*}{SciQ} & Accuracy & 94.07 & 97.09 & 96.88 & 96.68 & 95.38 & 91.06 \\
     & Confidence & 84.35 & 86.55 & 91.95 & 93.44 & 94.68 & 94.51 \\
     & ECE & 0.11 & 0.11 & 0.05 & 0.04 & 0.04 & 0.05 \\
     & (\%) Rounded & 99.80 & 98.99 & 99.40 & 78.39 & 97.59 & 97.29 \\
    \bottomrule
    \end{tabular}
    \caption{Performance of chat models across all question sets.}

\end{table*}

\clearpage

\begin{table*}[t]
    \centering

    \begin{tabular}{l|ccccc| c}
        \toprule
        \textbf{Provider}      &  \textbf{Google} & \textbf{OpenAI} & \textbf{Anthropic} & \textbf{Lambda} & \textbf{DeepSeek} & \textbf{Total}\\
        \midrule
        Spend (USD)   &  228.59   & 234.17 &  94.53    &   941.78 & $7.41^*$ & 1,506.48\\
        \bottomrule
        
    \end{tabular}
    \caption{Total spend by provider. New users on Google's platform receive $\$300$ in compute credits which we did not surpass. Lambda generously provided us with a $\$5,000$ research grant to cover our compute costs on their services.  We ran both Llama models using one NVIDIA H100 GPU over a combined 283 hours. Recording token distributions for later analysis significantly slowed down TPS. We suggest avoiding this when replicating our results. We did not keep track of our total spend on DeepSeek but we estimate the price based on publicly available pricing and our input and output token counts.} 
    \label{provider_spend}
\end{table*}

\begin{table*}[ht]
    \centering
    \small
    \begin{tabular}{l|lll}
        \toprule
        \textbf{Model}      & \textbf{Publisher}    & \textbf{Type}     & \textbf{Model Card} \\
        \midrule
        Claude Haiku 3      & Anthropic             & Chat              & \href{https://assets.anthropic.com/m/61e7d27f8c8f5919/original/Claude-3-Model-Card.pdf}{Model Card} \\
        Claude Sonnet 3.7   & Anthropic             & Reasoning         & \href{https://assets.anthropic.com/m/785e231869ea8b3b/original/claude-3-7-sonnet-system-card.pdf}{Model Card} \\
        Claude Sonnet 4     & Anthropic             & Reasoning         & \href{https://www-cdn.anthropic.com/6d8a8055020700718b0c49369f60816ba2a7c285.pdf}{Model Card} \\ 
        DeepSeek V3         & DeepSeek              & Chat              & \href{https://huggingface.co/deepseek-ai/DeepSeek-V3}{Model Card} \\
        DeepSeek R1         & DeepSeek              & Reasoning         & \href{https://huggingface.co/deepseek-ai/DeepSeek-R1}{Model Card} \\ 
        Gemini 2.5 Flash    & Google                & Chat              & \href{https://modelcards.withgoogle.com/assets/documents/gemini-2.5-flash.pdf}{Model Card} \\
        Gemini 2.5 Pro      & Google                & Reasoning         & \href{https://modelcards.withgoogle.com/assets/documents/gemini-2.5-pro.pdf}{Model Card} \\ 
        Llama 3.1 8B        & Meta                  & Chat              & \href{https://huggingface.co/meta-llama/Llama-3.1-8B-Instruct}{Model Card} \\
        Llama 3.1 70B       & Meta                  & Chat         & \href{https://huggingface.co/meta-llama/Llama-3.1-70B-Instruct}{Model Card} \\ 
        GPT 4o              & OpenAI                & Chat              & \href{https://openai.com/index/gpt-4o-system-card/}{Model Card} \\
        GPT o3              & OpenAI                & Reasoning         & \href{https://openai.com/index/o3-o4-mini-system-card/}{Model Card} \\
        \bottomrule
    \end{tabular}
    \caption{Information about each model used.}

\end{table*}

\clearpage

\end{document}